\documentclass[lettersize,journal]{IEEEtran}
\usepackage{amsmath,amsfonts,amssymb}
\usepackage{algorithmic}
\usepackage{algorithm}
\usepackage{array}
\usepackage{textcomp}
\usepackage{stfloats}
\usepackage{url}
\usepackage{verbatim}
\usepackage{graphicx}
\usepackage{xcolor}
\usepackage{tikz}
\usepackage{cite}
\usepackage{booktabs}
\usepackage[table]{xcolor}
\usepackage{graphicx}
\usepackage{colortbl}
\usepackage{amsmath}
\usepackage{multirow}
\usepackage{subfigure}
\usepackage{hyperref}
\usepackage{orcidlink}
\usepackage{xcolor}
\usepackage[normalem]{ulem} 
\usepackage{amsmath, amssymb}
\usepackage{makecell}
\newif\ifshowchanges
\showchangesfalse
\newcommand{\add}[1]{\ifshowchanges\textcolor{blue}{#1}\else #1\fi}
\newcommand{\del}[1]{\ifshowchanges\textcolor{red}{\sout{#1}}\fi}

\newcommand{\chg}[2]{\ifshowchanges\textcolor{blue}{#2}\textcolor{red}{\sout{#1}}\else #2\fi}
\newcommand{\showonly}[1]{\ifshowchanges #1\fi}

\usetikzlibrary{positioning}

\definecolor{best}{rgb}{0,0,0}
\definecolor{second}{rgb}{0,0,1}

\newcommand{\best}[1]{\textcolor{best}{\textbf{#1}}}
\newcommand{\second}[1]{\textcolor{best}{\underline{#1}}}
\begin{document}
\title{DiffFuSR: Super-Resolution of all Sentinel-2 Multispectral Bands using Diffusion Models}
\author{
Muhammad Sarmad\,\orcidlink{0000-0002-8635-9000}, 
Michael Kampffmeyer\,\orcidlink{0000-0002-7699-0405},~\IEEEmembership{Senior Member,~IEEE,} 
and Arnt-Børre Salberg\,\orcidlink{0000-0002-8113-8460},~\IEEEmembership{Member,~IEEE}

\thanks{Muhammad Sarmad and Arnt-Børre Salberg are with the Norwegian Computing Center, Oslo, Norway (e-mail: salberg@nr.no).}
\thanks{Michael Kampffmeyer is with the Department of Physics and Technology, UiT The Arctic University of Norway, Tromsø, Norway, and with the Norwegian Computing Center, Oslo, Norway.}

}

\markboth{preprint}%
{Shell \MakeLowercase{\textit{et al.}}: A Sample Article Using IEEEtran.cls for IEEE Journals}

\maketitle

\begin{abstract}

This paper presents \textit{DiffFuSR}, a modular pipeline for super-resolving all 12 spectral bands of Sentinel-2 Level-2A imagery to a unified ground sampling distance (GSD) of 2.5 meters. The pipeline comprises two stages: (i) a diffusion-based super-resolution (SR) model trained on high-resolution RGB imagery from the NAIP and WorldStrat datasets, harmonized to simulate Sentinel-2 characteristics; and (ii) a learned fusion network that upscales the remaining multispectral bands using the super-resolved RGB image as a spatial prior. We introduce a robust degradation model and contrastive degradation encoder to support blind SR. Extensive evaluations of the proposed SR pipeline on the OpenSR benchmark demonstrate that the proposed method outperforms current SOTA baselines in terms of reflectance fidelity, spectral consistency, spatial alignment, and hallucination suppression. Furthermore, the fusion network significantly outperforms classical \add{and learned} pansharpening approaches, enabling accurate enhancement of Sentinel-2's 20 m and 60 m bands. \del{This study underscores the power of harmonized learning with generative priors and fusion strategies to create a modular framework for Sentinel-2 SR.}\add{This work proposes a novel modular framework Sentinel-2 SR that utilizes harmonized learning with diffusion models and fusion strategies}. Our code and models can be found at \url{https://github.com/NorskRegnesentral/DiffFuSR}.

\end{abstract}


\section{Introduction}

\IEEEPARstart{T}{he} democratization of Earth observation has been significantly propelled by the advent of open-access satellite missions such as Sentinel-2, which offer freely available multispectral imagery on a global scale. This data serves as a valuable resource for a broad spectrum of applications, including land cover mapping, agricultural monitoring, urban planning, and disaster response~\cite{kansakar2016review}. \del{Despite the relatively high spatial resolution of some Sentinel-2 bands (up to 10 m), a substantial portion of the spectral channels, particularly those at 20 m and 60 m, remain too coarse for tasks that demand fine-grained spatial details. This disparity in resolution across bands limits the full exploitation of the data, especially in applications requiring pixel-level precision.} There exist several other satellite sensors that provide RGB and multispectral imagery with higher resolution than Sentinel-2, e.g., WorldView 2, 3 and Legion from Maxar, Pléiades from Airbus, and PlanetScope from Planet Labs. However, this data is not freely available, and the acquisition costs may be substantial. Given the global coverage and open availability of Sentinel-2's data, enhancing the spatial resolution through computational means is therefore a highly attractive prospect\add{~\cite{donike2025trustworthy,aybar2024sen2naip,10238735, kowaleczko2023real,rs12152366}}. 

\del{To fully leverage the potential of Sentinel-2's multispectral imagery, super-resolution (SR) has emerged as a compelling approach for enhancing the spatial details\mbox{~\cite{donike2025trustworthy,aybar2024sen2naip,10238735, kowaleczko2023real,rs12152366}}.}Super resolution (SR) refers to reconstructing high-resolution (HR) images from their lower-resolution counterparts, often using data-driven or model-based techniques. \del{Mathematically, SR is regarded as an ill-posed problem as there is a non-uniqueness of the solution. For a given low-resolution (LR) image, there are \del{theoretically an infinite number of}\add{numerous} HR images that could have produced that LR image when down-sampled. As the down-sampling process inherently loses high-frequency information (details), reversing this process means trying to infer this lost information, and many different "guesses" can be consistent with the LR input.}\add{SR is an inherently ill-posed problem, as multiple HR images can produce the same low-resolution (LR) counterpart, requiring models to infer lost high-frequency details.} Moreover, SR is also criticized for its reliance on learned priors and the risk of generating hallucinated details\mbox{~\cite{satlassuperres}}. However, when applied judiciously, with a clear understanding of its limitations, it can significantly improve both the visual interpretability and the utility of satellite imagery in downstream analytical tasks. 

In the context of Sentinel-2, which consists of multispectral bands in 10 m, 20 m and 60 m resolution, an opportunity arises to combine SR with data fusion techniques to enhance spatial resolution further. Unlike conventional SR, which relies solely on low-resolution inputs, fusion approaches, such as pansharpening, leverage the higher-resolution spectral bands (e.g., the 10 m RGB bands) as guiding information to reconstruct finer details in the lower-resolution bands \del{(e.g., 20 m and 60 m)}\mbox{~\cite{lanaras2018super,vivone2014critical}}. This reduces the ambiguity associated with SR by anchoring the reconstruction in real, higher-resolution measurements. 

\del{In this work, we propose DiffFuSR, a two-step pipeline that addresses the SR of Sentinel-2 imagery in a comprehensive manner. Our approach begins by super-resolving the 10 m-resolution RGB bands to an enhanced spatial resolution of 2.5 m. Focusing initially on RGB is particularly strategic: the majority of HR reference data available, whether from commercial satellites or aerial imagery, is limited to RGB channels, rather than the full 12-band Sentinel-2 spectrum. Consequently, prioritizing RGB SR allows for effective learning from widely available datasets while remaining consistent with practical downstream applications. 
Building upon the super-resolved 2.5 m RGB imagery, we then develop a data fusion strategy to upscale the remaining spectral bands to a unified 2.5 m spatial resolution. For each individual lower-resolution band, the corresponding 2.5 m RGB signal serves as a high-frequency spatial prior, guiding the fusion process. This design allows us to simultaneously preserve the spectral integrity of each band while enriching it with the fine-grained structural information captured in the RGB reference, thus mitigating the risks typically associated with hallucination in purely SR-based approaches.}

\add{While diffusion models have gained traction in natural image SR~\cite{rombach2022high, saharia2022image, teng2023relay}, their application to remote sensing SR remains limited. In this work, we propose DiffFuSR, a two-step pipeline that begins by super-resolving Sentinel-2's 10 m RGB bands to 2.5 m using a diffusion model trained on harmonized aerial and Sentinel-2 data. Subsequently, a learned fusion network upscales the remaining spectral bands, using the super-resolved RGB as a spatial prior to ensure spectral integrity and mitigate hallucination.}

\del{For the RGB SR step, we develop a diffusion-based model to enhance the spatial resolution of the Sentinel-2 RGB bands to 2.5 m. While diffusion models have gained traction in natural image generation\mbox{~\cite{rombach2022high, saharia2022image, teng2023relay}}, their application to remote sensing SR remains limited. In addition to diffusion models, }We leverage large and diverse training datasets and carefully address domain-specific challenges. We utilize a diverse corpus from aerial NAIP imagery~\cite{aybar2024sen2naip} and satellite-based WorldStrat data~\cite{cornebise2022open}, despite their substantial spectral and radiometric differences from Sentinel-2. To bridge these gaps, we apply spectral harmonization techniques~\cite{aybar2024sen2naip}, which prove critical for generalization. Additionally, we incorporate blind kernel modeling~\cite{wu2023conditional} to simulate unknown degradations typical in real-world Sentinel-2 data. Evaluations using the OpenSR benchmark~\cite{opensrtest} and qualitative results show that our models yield significant performance improvements over current state-of-the-art baselines~\cite{wolters2023zooming,cresson2022sr4rs,10887321,aybar2024sen2naip}.

To enhance the resolution of the remaining multispectral bands, \del{we design and compare three dedicated fusion pipelines targeting the 10 m, 20 m, and 60 m bands, respectively}\add{we design a neural network-based fusion module and train three instances of it for 10 m, 20 m, and 60 m bands, respectively}. Each fusion model leverages the super-resolved 2.5 m RGB imagery as a high-frequency spatial prior, adopting a structure inspired by pansharpening principles. To enable supervised learning without external HR references, we simulate LR training pairs using the Wald protocol~\cite{wald2002data} on native Sentinel-2 data. \del{The input configuration is tailored to each task: 10 m bands use only native 10 m inputs, ; 20 m bands combine 10 m and 20 m data; and 60 m bands integrate 10 m, 20 m, and 60 m channels.} This\del{targeted} design ensures that the fusion task fully exploits the available spatial information. Evaluations show that our learned fusion models significantly outperform classical \add{and learned baseline} methods \del{such as Gram-Schmidt pansharpening \mbox{\cite{dalla2015global}}} in terms of quantitative metrics such as ERGAS, PSNR, and SSIM, as well as improved visual fidelity and spectral consistency. 

In summary, \del{this work presents}\add{DiffFuSR is} a unified framework for super-resolving all Sentinel-2 bands to a common 2.5\,m resolution. \del{Unlike previous methods that have addressed either RGB SR or limited multispectral fusion separately, our approach integrates both into a modular, end-to-end pipeline capable of blind SR across all spatial resolutions and spectral channels.} The key contributions are summarized as follows:

\begin{itemize}

\item A novel diffusion-based SR framework for Sentinel-2 RGB bands, leveraging large-scale aerial and satellite datasets, enhanced with data harmonization and blind kernel modeling. \del{ to ensure robust generalization across diverse acquisition conditions.}

\item A new data fusion framework for multispectral SR that enhances the resolution of all Sentinel-2 bands (10 m, 20 m, and 60 m) to a unified 2.5 m spatial resolution using only native Sentinel-2 data, trained using Wald protocol simulations.

\item A comprehensive analysis of cross-domain SR challenges, demonstrating the critical importance of harmonization and degradation modeling through rigorous evaluation on the recently proposed OpenSR benchmark.

\end{itemize}

\del{The rest of the paper is organized as follows: Section II reviews existing SR models, with a focus on Earth observation applications. Section III details the methodologies that constitute the SR pipeline. Section IV describes the experiments and results, focusing on well-known benchmark datasets. The discussion is provided in Section VI, and finally, Section VII concludes this work.}

\section{Background and related works}

\del{Super-resolution (SR) has a rich history, with early breakthroughs focusing on enhancing the resolution by utilizing known relative displacements in image sequences \mbox{\cite{irani1991improving}}, and single image SR using of sparse representations in terms of coupled dictionaries jointly trained from high- and low-resolution image patch pairs \mbox{\cite{yang2010image}}. The introduction of deep learning in computer vision was early adapted for SR. Pioneering efforts that demonstrated the potential of convolutional neural network (CNNs) in enhancing image resolution include the SRCNN approach by Dong et al. \mbox{\cite{dong2015image}}.} 
\add{Early SR methods evolved from multi-image registration \cite{irani1991improving} and sparse representations \cite{yang2010image} to pioneering deep learning approaches like SRCNN \cite{dong2015image}, which established convolutional neural networks (CNNs) as the standard for the task.}

\del{Enhancing the resolution of remote sensing images, particularly from satellites, is extremely valuable, as it significantly enhances the utility of the data. Many satellite sensors (e.g., Landsat, WorldView 2 \& 3, SPOT 6, ...) have therefore been equipped with a panchromatic band in higher resolution, which offers a more detailed view of features on the ground.} Pansharpening, i.e., methods that utilize the panchromatic band \add{ available in many satellites (e.g., Landsat, WorldView etc.)} to enhance the resolution of simultaneously acquired multispectral data, has \del{consequently} been a research topic for decades~\cite{deng2022machine}. 


However, not all satellite sensors have a panchromatic band; hence, single-image SR has also become a central focus in remote sensing, particularly for enhancing the spatial resolution of Sentinel-2 imagery. Early work, such as Lanaras et al.~\cite{lanaras2018super}, focused on enhancing the resolution of the RGB bands using CNNs trained with synthetic data. Later methods expanded to multispectral SR through data fusion approaches, combining multiple Sentinel-2 revisits~\cite{10238735} or external HR sources such as PlanetScope~\cite{rs12152366}.

The generative power of generative adversarial networks (GANs) has also been explored for SR. Early works include the SRGAN approach by Ledig et al.~\cite{ledig2017photo}, which were further \del{enhanced}\add{improved} by Wang et al.~\cite{wang2018esrgan}. The benefit of using generative methods like GANs for SR is the potential of creating photo-realistic images, contrary to non-generative methods, which tend to create too smooth results~\cite{ledig2017photo}. \del{After gaining influence in computer vision, GANs were also adopted to enhance the resolution of remote sensing images (see e.g.,\mbox{~\cite{ma2018super}}).} Recent advances in generative modeling have motivated the exploration of diffusion models for SR tasks~\cite{saharia2022image,rombach2022high}. \del{In the remote sensing domain, }Wu et al.\mbox{~\cite{wu2023conditional}} applied conditional diffusion models for Sentinel-2 SR, enhancing only the RGB bands by leveraging paired SPOT 6/7 imagery. However, their approach remains constrained to RGB bands, and the reliance on paired HR datasets limits its direct scalability to the remaining Sentinel-2 bands.



Efforts such as the WorldStrat dataset~\cite{cornebise2022open} and the OpenSR benchmark~\cite{opensrtest}, have provided valuable resources for training and evaluating remote sensing SR models under real-world conditions. Wolters et al.~\cite{satlassuperres} leveraged NAIP aerial imagery for generative SR, but faced challenges due to spectral and radiometric inconsistencies between NAIP and Sentinel-2, leading to domain gaps that degraded the performance. The SEN2NAIP dataset and accompanying work by Aybar et al.~\cite{aybar2024sen2naip} introduced a harmonized degradation model to better align NAIP images with Sentinel-2 characteristics, proving crucial for cross-sensor learning. Donike et al.~\cite{donike2025trustworthy} employed latent diffusion models trained on SEN2NAIP, demonstrating that operating in the latent space can significantly reduce resource consumption. They focus solely on super-resolving the 10 m Sentinel-2 bands.

\del{Beyond SR, pansharpening techniques have historically been used to fuse HR spatial information with lower-resolution multispectral data} \add{After performing SR on the 10 m bands, pansharpening techniques can be applied to fuse the super-resolved image  with the lower-resolution multispectral bands (e.g., 20 m and 60 m)}~\cite{sarmad2024diffusion}. Classical methods such as Gram-Schmidt pansharpening~\cite{dalla2015global} can be effective but may prove insufficient or introduce spectral distortions when attempting to fuse super-resolved RGB images (acting as a pseudo-panchromatic band) with Sentinel-2's remaining multispectral bands, especially when targeting a significant resolution enhancement to 2.5 m \cite{sarmad2024diffusion}. Lanaras et al.~\cite{lanaras2018super} demonstrated that neural networks can perform such fusion more effectively than classical methods, mitigating some distortions. However, their method was limited to enhancing bands to a 10 m ground sampling distance (GSD) \add{instead of a higher resolution of 2.5 m GSD}.

The modular framework presented in this paper significantly extends previous work by jointly addressing SR of all Sentinel-2 bands to a unified 2.5\,m resolution. First, we super-resolve the RGB bands using an efficient diffusion model trained on harmonized data, ensuring robust generalization. Second, a learned fusion network, inspired by pansharpening principles but trained for superior spectral fidelity, upscales the remaining 10\,m, 20\,m, and 60\,m Sentinel-2 bands. While prior works address isolated components of this problem, to the best of our knowledge, this is the first unified approach that performs blind SR across all Sentinel-2 bands and resolutions.

\section{Method}

\begin{figure}[t]
    \centering
    \includegraphics[width=0.45\textwidth]{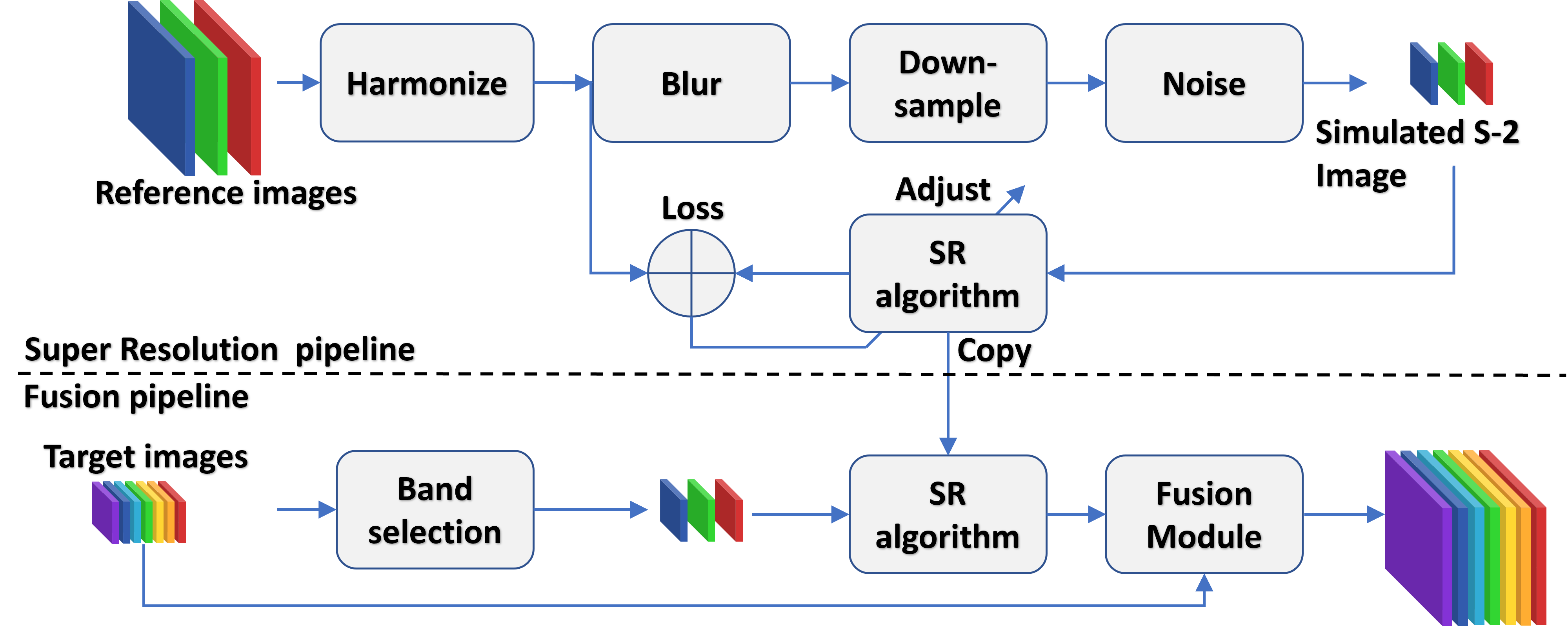}
    \caption{Overview of the proposed pipeline combining RGB SR and multispectral fusion. The top branch illustrates the training of a denoising diffusion model using synthetic low-resolution RGB images generated through harmonization, blurring, downsampling, and noise addition. The model learns to reconstruct HR RGB images. The bottom branch depicts the fusion process: multispectral Sentinel-2 bands are fused with the super-resolved RGB output via a learned fusion module to produce full-resolution 12-band output.}
    \label{fig:superai_pipeline}
\end{figure}

\del{The goal of this work is to super-resolve all 12 Sentinel-2 multispectral bands to a unified 2.5\,m GSD.} First, in section~\ref{sec:sr}, we describe a conditional denoising diffusion model to super-resolve the RGB bands to 2.5\,m GSD, trained on harmonized synthetic data to ensure robust generalization. Secondly, in section~\ref{sec:fusion}, we introduce a learned multispectral fusion network that leverages the super-resolved RGB output to upscale the remaining Sentinel-2 bands to the target GSD of 2.5\,m. The overall pipeline is illustrated in Fig.~\ref{fig:superai_pipeline}. \del{Next, we will describe each module of the pipeline in detail.}

\subsection{Super-Resolution of RGB bands using Conditional Denoising Diffusion Models}
\label{sec:sr}
\begin{figure}[t]
\centering
\includegraphics[width=0.48\textwidth]{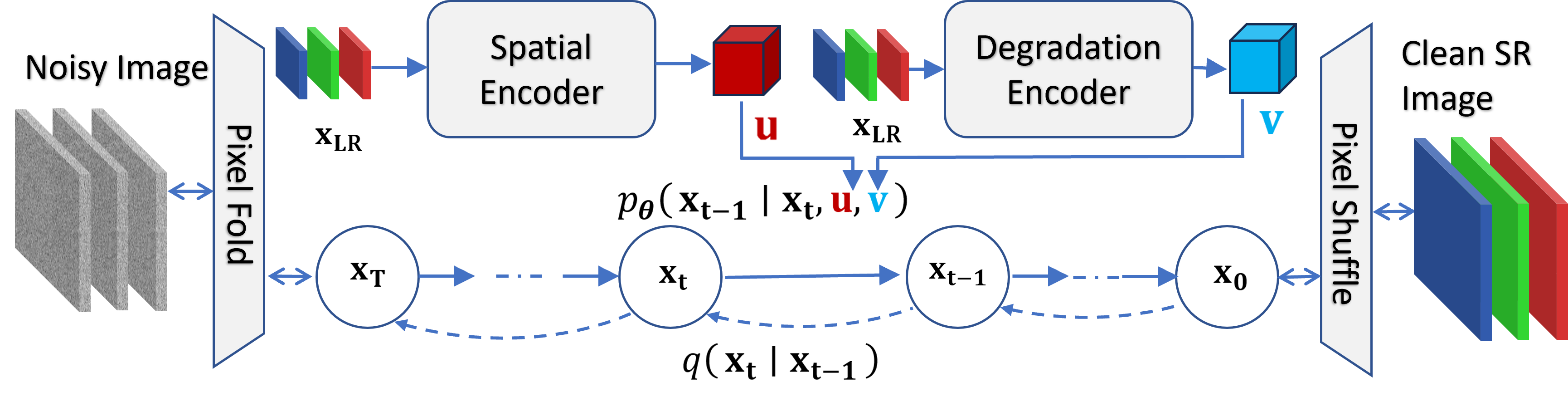}
\caption{Architecture of the proposed conditional denoising diffusion model for RGB SR. The model learns to reconstruct a HR RGB image \( \mathbf{x}_0 \) from a noisy latent \( \mathbf{x}_T \) via a denoising process modeled as a Markov chain. The input image is first degraded with Gaussian noise and spatially rearranged using a pixel-folding operation. A spatial encoder extracts contextual features \( \mathbf{u} \) from the low-resolution input, while a degradation encoder produces a global degradation embedding \( \mathbf{v} \). Both \( \mathbf{u} \) and \( \mathbf{v} \) are used to condition the denoising network at each reverse time step \( t \). The sequence \( \mathbf{x}_T \rightarrow \dots \rightarrow \mathbf{x}_0 \) is denoised through a learned reverse diffusion process. The final output is reshaped using a pixel shuffle operation to recover the super-resolved RGB image.}
\label{fig:diffusion_pipeline}
\end{figure}

We formulate the RGB SR task as conditional generation using a denoising diffusion probabilistic model (DDPM) inspired from \cite{wu2023conditional}. The goal is to generate a HR RGB image \( \mathbf{x}_0 \in \mathbb{R}^{C \times H \times W} \), conditioned on a LR observation \( \mathbf{x}_{\text{LR}} \in \mathbb{R}^{C \times H' \times W'} \), where \( H' < H \) and \( W' < W \). The model learns to reverse a Gaussian noise corruption process added to \( \mathbf{x}_0 \), using the fixed \( \mathbf{x}_{\text{LR}} \) to create multiple conditioning inputs. The SR pipeline consists of a forward diffusion process and a reverse denoising process (Fig.~\ref{fig:diffusion_pipeline}).

\subsubsection{Forward Diffusion Process}

Following \cite{ho2020denoising}, the forward process is a Markov chain that progressively adds Gaussian noise to a "clean" input image \( \mathbf{x}_0 \) over \( T \) steps:
\begin{equation}
    q(\mathbf{x}_t | \mathbf{x}_{t-1}) = \mathcal{N}(\mathbf{x}_t; \sqrt{\alpha_t} \mathbf{x}_{t-1}, (1 - \alpha_t)\mathbf{I}),
\end{equation}
with a predefined noise schedule \( \{\alpha_t\}_{t=1}^T \). This defines the marginal distribution:
\begin{equation}
    q(\mathbf{x}_t | \mathbf{x}_0) = \mathcal{N}(\mathbf{x}_t; \sqrt{\bar{\alpha}_t} \mathbf{x}_0, (1 - \bar{\alpha}_t)\mathbf{I}),
\end{equation}
where \( \bar{\alpha}_t = \prod_{s=1}^t \alpha_s \). At \( t = T \), the sample \( \mathbf{x}_T \) approximates a standard Gaussian.

\subsubsection{Reverse Denoising Process}
The reverse denoising process is modeled as a Markov chain conditioned on the LR input \( \mathbf{x}_{\text{LR}} \). 
Starting from Gaussian noise \( \mathbf{x}_T \sim \mathcal{N}(\mathbf{0}, \mathbf{I}) \), 
this process iteratively denoises a sequence of latent variables \( \mathbf{x}_T, \dots, \mathbf{x}_0 \) according to:
\begin{equation}
    p_{\boldsymbol{\theta}}(\mathbf{x}_{0:T} | \mathbf{u}, \mathbf{v}) 
    = p(\mathbf{x}_T)\prod_{t=1}^{T} p_{\boldsymbol{\theta}}(\mathbf{x}_{t-1} | \mathbf{x}_t, \mathbf{u}, \mathbf{v}),
\end{equation}
where each reverse transition \( p_{\boldsymbol{\theta}}(\mathbf{x}_{t-1} | \mathbf{x}_t, \mathbf{u}, \mathbf{v}) \) is modeled as a Gaussian distribution:
\begin{equation}
    p_{\boldsymbol{\theta}}(\mathbf{x}_{t-1} | \mathbf{x}_t, \mathbf{u}, \mathbf{v}) 
    = \mathcal{N}(\mathbf{x}_{t-1}; \boldsymbol{\mu}_{\boldsymbol{\theta}}(\mathbf{x}_t, t, \mathbf{u}, \mathbf{v}), \sigma_t^2 \mathbf{I}).
\end{equation}
The variance \( \sigma_t^2 \) is typically pre-defined, while the mean \( \boldsymbol{\mu}_{\boldsymbol{\theta}} \) is predicted by a denoising network, commonly a U-Net architecture~\cite{ronneberger2015u}, conditioned on the noisy latent \( \mathbf{x}_t \), the timestep \( t \), and the fixed context \( \mathbf{u}, \mathbf{v} \). 
\add{The integration of \( \mathbf{u} \) and \( \mathbf{v} \) allows the model to incorporate both spatial structural details and degradation-specific behavior, thereby enabling accurate and blind SR.}

The conditioning inputs \( \mathbf{u} \) and \( \mathbf{v} \) are fixed throughout the reverse process and derived from the observed LR image \( \mathbf{x}_{\text{LR}} \) as follows:

\begin{itemize}
    \item The spatial feature map \( \mathbf{u} = f_{\boldsymbol{\phi}}(\mathbf{x}_{\text{LR}}) \) is extracted by an encoder \( f_{\boldsymbol{\phi}} \). It captures coarse structural information, such as edges and object geometries from \( \mathbf{x}_{\text{LR}} \). The spatial feature map, \( \mathbf{u} \), is concatenated at the input of the UNet. 
    
    \item The global degradation embedding \( \mathbf{v} = g_{\boldsymbol{\psi}}(\mathbf{x}_{\text{LR}}) \) summarizes the underlying blur, noise, and downsampling characteristics of \( \mathbf{x}_{\text{LR}} \). This embedding is computed using a separate encoder $ g_{\boldsymbol{\psi}} $, trained via contrastive learning~\cite{he2020momentum}\del{, and serves to modulate the denoising network through degradation-aware convolutions. The degradation embedding is injected at multiple positions in the network since it modulates the kernels of degradation-aware convolution layers}.
\end{itemize}
\del{The integration of \( \mathbf{u} \) and \( \mathbf{v} \) allows the model to incorporate both spatial structural details and degradation-specific behavior, thereby enabling accurate and blind SR.}

\subsubsection{Network Architecture}

The proposed diffusion model is comprised of three main components (Fig.~\ref{fig:diffusion_pipeline}): (i) a \textit{denoising U-Net} that predicts the clean HR image from noisy input, (ii) a \textit{spatial encoder} that extracts contextual features from the low-resolution image, and (iii) a \textit{degradation encoder} trained with contrastive learning to summarize the degradation characteristics. We follow the design by Wu et al.~\cite{wu2023conditional}:

\paragraph{Denoising U-Net}

The core network \( h_{\boldsymbol{\theta}} \) is a conditional U-Net that predicts the clean image \( \mathbf{x}_0 \) given a noisy input \( \mathbf{x}_t \), timestep \( t \), and conditioning vectors \( \mathbf{u}, \mathbf{v} \). To improve computational efficiency without losing information, we apply a \textit{pixel folding} operation to \( \mathbf{x}_t \) at the input~\cite{wu2023conditional}. This reduces spatial resolution by a factor \( r \) while increasing the number of channels by \( r^2 \). Formally, this maps an image of dimension \( C \times H \times W \) to \( (C \cdot r^2) \times (H / r) \times (W / r) \). \del{The inverse of this operation is pixel shuffle as proposed in\mbox{~\cite{shi2016real}}. We use pixel shuffle to retrieve the full-sized final image.}
The folded input is passed through the \add{denoising} U-Net~\cite{wu2023conditional}.\del{, which consists of:}
\ifshowchanges

\begin{itemize}
    \item .\del{a downsampling path with multiple feature extraction blocks,}
    \item .\del{an upsampling path with nearest-neighbour interpolation and convolution (i.e., spatial resolution is increased by replicating neighboring pixels, followed by refinement with convolution layers),}
    \item .\del{residual blocks that incorporate conditioning via timestep embedding and degradation information.}
\end{itemize}
\else
\fi
After denoising, the output is unfolded back to the original spatial resolution using the pixel shuffle operation~\cite{shi2016real}.


To inform the network of the current step in the reverse denoising process, a timestep embedding is injected at each residual block. The scalar timestep \( t \in \{1, \dots, T\} \) is first mapped to a sinusoidal positional encoding:
\[
\boldsymbol{\kappa}(t) = [\sin(\omega_1 t), \cos(\omega_1 t), \dots, \sin(\omega_d t), \cos(\omega_d t)]^T,
\]
and then passed through a multi-layer perceptron (MLP) to produce the embedding \( \mathbf{t}_e \). This embedding is added to intermediate feature maps throughout the U-Net~\cite{wu2023conditional}.

\paragraph{Spatial Encoder}

The LR input \( \mathbf{x}_{\text{LR}} \) is processed by a spatial encoder \( f_{\boldsymbol{\phi}} \), implemented as an RRDB-Net consisting of 23 Residual-in-Residual Dense Blocks (RRDBs) with 64 feature channels\del{. This architecture is known for its strong performance in SR tasks, particularly its ability to extract hierarchical spatial features without normalization layers }~\cite{wang2018esrgan}. The encoder outputs a spatial feature map \( \mathbf{u} = f_{\boldsymbol{\phi}}(\mathbf{x}_{\text{LR}}) \), which serves as a global spatial conditioning signal for the denoising model.

Rather than being fused at multiple layers, the spatial feature map \( \mathbf{u} \) is concatenated with the input of the U-Net denoiser at the first layer only. This provides the network with high-level spatial context from the LR image while keeping the fusion interface lightweight. During training, the decoder (a light weight CNN, not shown in Fig.~\ref{fig:diffusion_pipeline}) is attached to the encoder to predict a coarse HR reconstruction \( \widehat{\mathbf{x}}_0 \), supervised with an \( \ell_1 \) consistency loss. The decoder is discarded during inference.

\paragraph{Degradation Encoder}

To model the implicit degradation of the LR image, we use a lightweight CNN \( g_{\boldsymbol{\psi}} \) that maps an LR image patch \( \mathbf{x}_{\text{LR}} \) to a global degradation vector \( \mathbf{v} = g_{\boldsymbol{\psi}}(\mathbf{x}_{\text{LR}}) \)\del{, following the contrastive learning formulation of\mbox{~\cite{wang2021unsupervised}}}. The network consists of several convolutional layers, followed by global average pooling and a projection head. It is trained with a contrastive objective\mbox{~\cite{wang2021unsupervised}}, \add{where}\del{. In the contrastive setting,} the goal is to train a model such that given two crops \( \mathbf{x}_q \), \( \mathbf{x}_k \) from the same LR image (i.e., with the same degradation), their embeddings \( \mathbf{v}_q \), \( \mathbf{v}_k \) should be similar, while embeddings from different images (assumed to have different degradations) should be dissimilar. Under this setting, the model learns about the degradation in an unsupervised manner.

The learned degradation vector \( \mathbf{v} \) is injected into each residual block in the U-Net using a degradation-aware convolution (DAConv) layer\del{, as proposed by Wu et al.}~\cite{wu2023conditional}\del{, which adapts the convolutional filters based on the conditioning vector}. In this approach, the convolutional filters are dynamically modulated based on the conditioning vector, allowing the network to adapt to different degradation types during inference.

\subsubsection{Training Objective}

The model is trained end-to-end using a combination of three loss terms:

\begin{itemize}
    \item \textit{Denoising loss:} The diffusion model is trained to predict the clean HR image \( \mathbf{x}_0 \) from its noisy version \( \mathbf{x}_t \), conditioned on spatial and degradation features using the ELBO loss~\cite{ho2020denoising}:
    \begin{equation}
        \mathcal{L}_{\text{elbo}} = \left\| \mathbf{x}_0 - h_{\boldsymbol{\theta}}(\mathbf{x}_t, t, \mathbf{u}, \mathbf{v}) \right\|_1.
    \end{equation}

\item \textit{Consistency loss:} An auxiliary decoder, \(\xi(\cdot)\), is attached to the spatial encoder \( f_{\boldsymbol{\phi}} \), which processes the LR input \( \mathbf{x}_{\text{LR}} \). The decoder predicts a coarse HR estimate \( \widehat{\mathbf{x}}_0 = \xi(f_{\boldsymbol{\phi}}(\mathbf{x}_{\text{LR}})) \), and the output is compared against the ground truth \( \mathbf{x}_0 \) using an L1 loss:
\begin{equation}
    \mathcal{L}_{\text{consis}} = \left\| \widehat{\mathbf{x}}_0 - \mathbf{x}_0 \right\|_1.
\end{equation}
This encourages the spatial encoder to learn features that are aligned with the HR target. The decoder is only used during training and discarded at inference.

\item \textit{Contrastive loss:} 
We use the InfoNCE loss for training~\cite{wu2023conditional}. Given two patches \( \mathbf{x}_{\text{LR}} \) and \( \mathbf{x}_{\text{LR}}^+ \) from the same image (positive pair), and a set of negatives \( \{\mathbf{x}_{\text{LR}}^- \} \) from different images in the batch, the contrastive loss is defined as:

\begin{equation}
\mathcal{L}_{\text{contrast}} = -\log \frac{\exp(\text{sim}(\mathbf{v}, \mathbf{v}^+)/\tau)}{\sum\limits_{j} \exp(\text{sim}(\mathbf{v}, \mathbf{v}_j)/\tau)},
\end{equation}

where \( \mathbf{v} = g_{\boldsymbol{\psi}}(\mathbf{x}_{\text{LR}}) \), \( \mathbf{v}^+ = g_{\boldsymbol{\psi}}(\mathbf{x}_{\text{LR}}^+) \), \( \mathbf{v}_j \) includes all negative samples, \( \tau \) is a temperature hyperparameter, and \( \text{sim}(\cdot, \cdot) \) denotes cosine similarity, i.e., \( \text{sim}(\mathbf{a}, \mathbf{b}) = \frac{\mathbf{a}^T \mathbf{b}}{\|\mathbf{a}\| \|\mathbf{b}\|} \).

\end{itemize}
The total training loss is then a weighted sum:
\begin{equation}
    \mathcal{L}_{\text{total}} = \mathcal{L}_{\text{elbo}} + \mathcal{L}_{\text{consis}} + \lambda_{\text{contrast}} \cdot \mathcal{L}_{\text{contrast}},
\end{equation}
where the constant \(\lambda_{\text{contrast}}\) is a weighting factor that determines the contribution of the contrastive loss \(\mathcal{L}_{\text{contrast}}\) relative to the ELBO loss \(\mathcal{L}_{\text{elbo}}\) and the consistency loss \(\mathcal{L}_{\text{consis}}\). A small default value, such as \(\lambda_{\text{contrast}} = 0.01\), ensures that the contrastive objective improves representation learning without overwhelming the main generative and consistency objectives.

\subsubsection{Sampling}

At inference time, a HR image is sampled by starting from \( \mathbf{x}_T \sim \mathcal{N}(0, \mathbf{I}) \) and iteratively applying the learned denoising model conditioned on \( \mathbf{u} \) and \( \mathbf{v} \). The final output \( \mathbf{x}_0 \) is returned as the super-resolved RGB image.

\del{Both the spatial encoder \( f_{\boldsymbol{\phi}} \) and degradation encoder \( g_{\boldsymbol{\psi}} \) receive the same degraded image as input. The randomized degradation ensures that the degradation vector \( \mathbf{v} \) is learned from data without relying on ground truth kernels which is key component of the model's blind SR capability. Next, we will describe how we degrade and prepare the HR images for training the diffusion model.}

\subsubsection{Simulating Sentinel-2 images}

\del{Due to the absence of HR Sentinel-2 imagery suitable for supervised learning, we use HR NAIP aerial images as a proxy to construct synthetic training pairs.}

\del{Previous works such as the SEN2NAIP study\mbox{~\cite{aybar2024sen2naip}} leverage real, co-registered NAIP–Sentinel-2 pairs to train a harmonization model, comparing a deterministic neural network-based approach with a statistical method grounded in power-law transformations. While the authors did not claim a definitive advantage for either method, they noted that the neural network approach produced slightly sharper visual results for SR.} 
\del{In contrast, we found that learned harmonization does not consistently yield sharper outputs and introduces unnecessary computational overhead. Therefore, we adopt a simpler, statistics-based harmonization model using per-channel power-law transformations.}

\add{We adopt a simple, statistics-based harmonization model using per-channel power-law transformations to generate synthetic Sentinel-2-like images from NAIP images. This is achieved by a two-stage transformation:}
\del{We generate synthetic LR Sentinel-2-like images by applying a two-stage transformation to NAIP data:} (i) spectral harmonization to match Sentinel-2 reflectance distributions, and (ii) randomized degradation to simulate optical and sensor effects such as blur and noise. This strategy enables fully self-supervised training and supports blind SR without reliance on paired ground truth.

\paragraph{Harmonization Model}

Harmonization adjusts the spectral distribution of NAIP imagery to resemble the Sentinel-2 RGB bands\del{. This process mitigates radiometric discrepancies between the two sensors}, enabling more effective cross-domain learning. We perform per-channel gamma correction using a power-law transformation:

\begin{equation}
    \widehat{\mathbf{x}}^c_{NP} = \left(\frac{\mathbf{x}^c_{NP}}{k} \right)^{\frac{1}{\gamma_c}} \cdot k,
\end{equation}
where \(\mathbf{x}^c_{NP}\)  denotes band \( c \) of the NAIP image, \( k = 255 \) denotes the dynamic range, and \( \gamma_c \) is the channel-specific gamma value. These gamma values are \del{not arbitrarily chosen; rather, they are }derived empirically by aligning NAIP reflectance distributions to co-registered Sentinel-2 imagery. \del{Following the methodology of Aybar et al.\mbox{~\cite{aybar2024sen2naip}}, }Optimal gamma values are computed for each channel across a large corpus of image pairs~\cite{aybar2024sen2naip}. A multivariate normal model is then fit to the resulting gamma vectors, and the median values are used to define a global correction profile. This ensures consistent spectral alignment during inference without the need for image-specific tuning.

\paragraph{Degradation Model}

To simulate Sentinel-2-like degradations, we apply a blind degradation strategy that combines randomized blur, downsampling, and additive noise. The degraded image is modeled as:

\begin{equation}
    \widehat{\mathbf{x}}_{S2} = \left[\Phi_{\boldsymbol{\rho}} (\widehat{\mathbf{x}}_{NP} ) \right]_{\downarrow} + \mathbf{n}, 
\end{equation}
where \( \Phi_{\boldsymbol{\rho}}(\cdot ) \) is a Gaussian blur operator parameterized by \( \boldsymbol{\rho} \), \( \downarrow \) denotes bicubic downsampling, and \( \mathbf{n} \) is additive Gaussian noise. 

\paragraph{Blur Model}

To emulate optical blur, we use both isotropic and anisotropic Gaussian kernels, with parameters sampled randomly for each training image. We select a fixed kernel size of \( 21 \times 21 \). During training, we use an anisotropic Gaussian blur type, whereas for validation, we use an isotropic Gaussian. We sample the smoothing parameter \( \sigma \)  from the range \(\sigma \in [2.0, 4.0]\), the eigenvalue range of the anisotropic Gaussian kernel as \( \lambda_1, \lambda_2 \in [0.2, 4.0] \), and the rotation angle as 
 \( \theta \in [0, \pi] \). \del{Our strategy is different from Aybar et al. \mbox{\cite{aybar2024sen2naip}} since they suggested applying a fixed smoothing parameter (\(\sigma\)) for the blur kernel. From empirical evaluations, we concluded that a blind approach led to better results. In addition, we also found that using both isotropic and anisotropic blur is important.}

\paragraph{Noise Model}

Sensor noise is modeled as additive Gaussian noise, i.e.,
\begin{equation}
    \mathbf{n} \sim \mathcal{N}(0, \sigma_n^2\mathbf{I}), \quad \sigma_n \in [0.0, 25.0],
\end{equation}
with \( \sigma_n \) sampled randomly for each image during training.

\paragraph{Blind Conditioning}

The same degraded image is passed to both the spatial encoder \( f_{\boldsymbol{\phi}} \) and the degradation encoder \( g_{\boldsymbol{\psi}} \). The randomized, unsupervised nature of the degradation process enables the model to generalize to unknown test-time corruptions. In particular, the degradation representation \( \mathbf{v} \) is learned entirely without supervision of degradation parameters, using contrastive learning to capture differences in blur, noise, and downsampling implicitly.

\subsection{Multispectral Band Fusion}
\label{sec:fusion}

\begin{figure}[t]
    \centering
    \centering
    \includegraphics[width=0.480\textwidth]{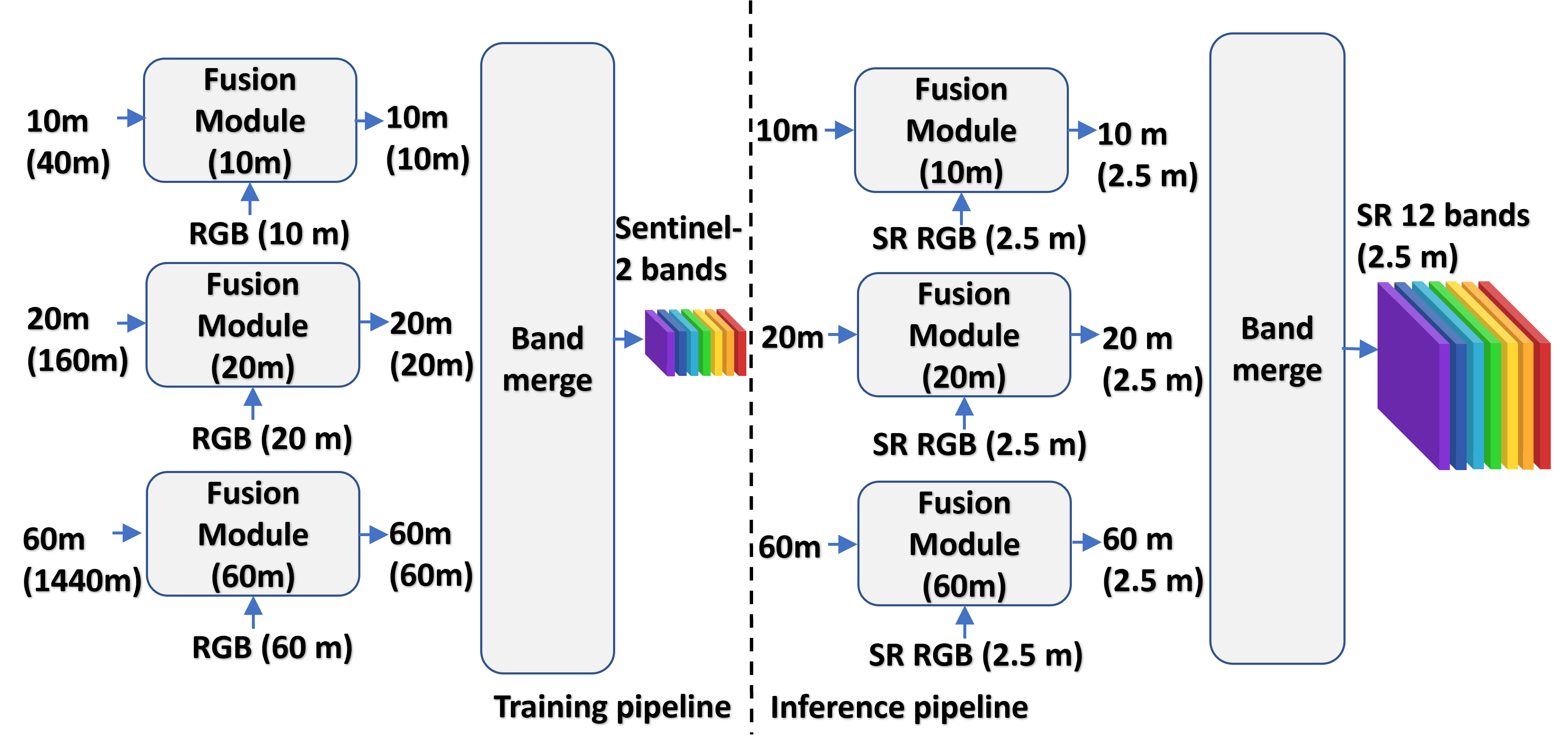}
    \caption{Architecture of the multispectral fusion pipeline. At test time, each resolution group of 10 m, 20 m, and 60 m bands is processed by a dedicated fusion module that receives the native-resolution multispectral inputs along with the RGB image super-resolved to 2.5 m. These modules independently reconstruct each group at 2.5 m GSD. The outputs are then merged to produce a consistent 12-band super-resolved Sentinel-2 image. This process is applied during both training (using synthetic degraded inputs) and inference (using native-resolution bands), with shared architecture and supervision strategy across all groups. The scaling factor for degraded input for training is computed using the Wald protocol. \del{For example, for the 60 m group, we aim to learn the mapping from 60 m to 2.5 m given a 2.5 m SR guidance signal. This corresponds to a 24× scale factor. During training, all bands (10 m, 20 m, and 60 m) are degraded by this factor, producing input resolutions of 240 m, 480 m, and 1440 m, respectively.} \add{For example, for the 60 m group, we aim to learn the mapping from 60 m to 2.5 m given a 2.5 m SR guidance signal. This corresponds to a 24× scale factor. During training, 60m bands are degraded by this factor, producing input resolutions of 1440 m.} At training time, the network reconstructs the degraded signal to 60 m GSD, supervised by the corresponding 60 m ground-truth reference.}
    \label{fig:fusion_pipeline}
\end{figure}

\begin{figure}[t]
    \centering
    \includegraphics[width=0.48\textwidth]{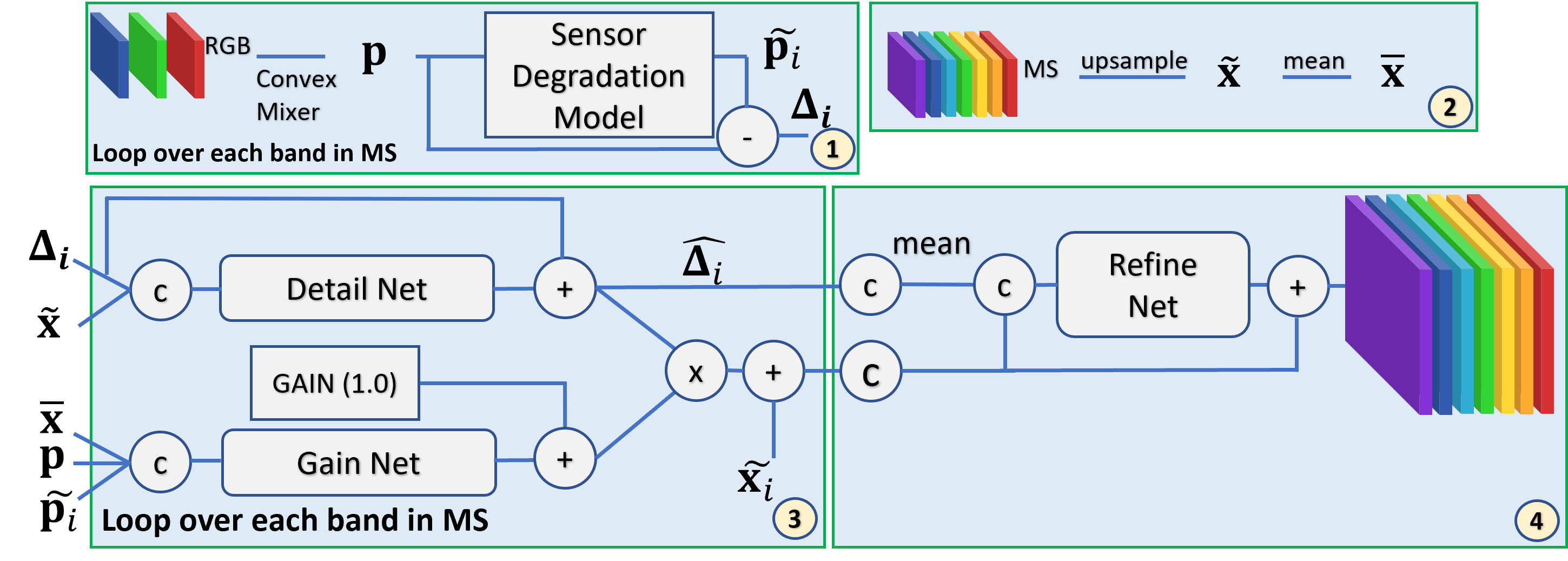}
    \caption{\add{Architecture of the proposed GLP-inspired \cite{vivone2018full} fusion module. Each model variant for the 10\,m, 20\,m, and 60\,m branches contains approximately 202k trainable parameters. The symbols denote basic operations: ``+'' indicates element-wise addition, ``$-$'' indicates subtraction, ``$\times$'' indicates element-wise multiplication, and ``c'' indicates channel-wise concatenation.}}

    \label{fig:glpnn_pipeline}
\end{figure}
    
To produce a consistent 2.5 m resolution across all Sentinel-2 bands, we design a learnable fusion approach that reconstructs HR multispectral bands by injecting spatial details from the super-resolved RGB image, $\mathbf{x}_0$ (Fig.~\ref{fig:fusion_pipeline}). 
This is formulated as a guided upsampling problem, \del{increasing resolution by pixel replication and feature refinement} where the RGB image provides high-frequency structural cues and the LR multispectral bands contribute the spectral content.

Let \( \mathbf{X}_{\text{LR}} \) denote a multispectral data at its native resolution (10 m, 20 m, and 60 m). The  objective of the fusion module, \(F_{\boldsymbol{\nu}}(\cdot, \cdot)\), is to learn a mapping:
\[
\widehat{\mathbf{X}}_{\text{HR}} = F_{\boldsymbol{\nu}}(\mathbf{X}_{\text{LR}}, \mathbf{x}_{0}),
\]
\add{where \( \widehat{\mathbf{X}}_{\text{HR}} \) is the reconstructed data at 2.5m GSD, and}
\del{such that the reconstructed data \( \widehat{\mathbf{X}}_{\text{HR}} \) captures fine structural details in with the super-resolved Sentinel-2 RGB image at 2.5 m GSD,} \del{from the diffusion module} \del{ \( \mathbf{x}_{0} \), while remaining spectrally consistent with the original \( \mathbf{X}_{\text{LR}} \) data. Here} \( \boldsymbol{\nu}\) denotes the parameters of the fusion model. In practice, the fusion pipeline contains three different modules, one for each band group (10m, 20m, 60m), which are trained and tested simultaneously but completely independently of each other (Fig.~\ref{fig:fusion_pipeline}).

\add{The three proposed fusion pipelines are based on the Generalized Laplacian Pyramid (GLP) with modulation transfer function (MTF) filters and a regression injection model (MTF-GLP) \cite{vivone2018full, aiazzi2002context}, as illustrated in Fig.~\ref{fig:glpnn_pipeline}. First, the panchromatic image is low-pass filtered by band-specific MTF filters, then down- and up-sampled. Subtracting this result from the original panchromatic image yields the spatial details for injection. The injection coefficients, which control how much detail is added to each multispectral band, are estimated by a regression procedure.}

\add{
We embed the main components of MTF-GLP into the neural network architecture. The per-band low-pass filtering is implemented as fixed convolutional filters mimicking the sensor’s MTF, corresponding to the panchromatic degradation step in the classical model. GLP-style detail extraction is realized by subtracting the low-pass filtered panchromatic image from the original panchromatic image, with a residual detail network learning only the correction beyond the classical detail term. The regression-based injection mechanism is a gain estimator, initialized around unity, which adapts the detail injection for each band. Finally, a shallow refinement network enforces inter-band consistency, compensating for nonlinear effects not captured by linear regression.}

\subsubsection{Training Strategy and Degradation Simulation}

The fusion model is trained in a fully self-supervised manner using only native Sentinel-2 data. \del{No HR targets are assumed. Instead,} We adopt \add{a Wald protocol} \del{a synthetic} training strategy where the input bands are downsampled from their native resolution, and the original band serves as the supervision target. \del{This follows the Wald protocol in spirit, though adapted for groups with no native HR ground truth (i.e., 20 m and 60 m bands).}

To simulate realistic \del{LR observations}\add{downsampled images} during training, each input band undergoes a degradation process \del{comprising}
\add{
\begin{equation}
    \mathbf{X}_{\text{LR}}^{\downarrow} = 
    \left[\Phi_{\boldsymbol{\sigma}}
    \left(
    \mathbf{X}_{\text{LR}}
    \right)\right]_\downarrow,
\end{equation}
where $\Phi_{\boldsymbol{\sigma}}(\cdot )$ is an optional isotropic Gaussian blur is applied to emulate sensor response, and $\downarrow$ denotes the boxcar filter, implemented as average pooling with stride corresponding to the target scale.
}

\ifshowchanges

\begin{itemize}
    \item \textit{\del{Blur}}\textbf{:} \del{An optional isotropic Gaussian blur is applied to emulate sensor response.}
    \item \textit{\del{Downsampling}}\textbf{:} \del{Spatial resolution is reduced using a \textit{boxcar filter}, implemented as average pooling with stride corresponding to the target scale. This approximates spatial averaging seen in optical sensor acquisition.}
\end{itemize}

\else
\fi

\del{At test time, the fusion model receives inputs at the original resolution (10 m, 20 m, or 60 m), along with the super-resolved RGB image in  2.5 m resolution from the diffusion model. The super-resolved RGB input is not used during training. However, the network is expected to learn to generalize zero-shot to enhanced spatial priors from the super-resolved RGB image at 2.5 m\mbox{~\cite {lanaras2018super}}.}


\subsubsection{Network Implementation}

\chg{The fusion model  \( F_{\boldsymbol{\nu}}(\cdot, \cdot) \) is a separate model for each resolution group (10 m, 20 m, and 60 m), each implemented as a two-branch convolutional network. One branch processes the degraded multispectral inputs, while the other processes a fusion signal derived from auxiliary inputs (e.g., RGB and NIR bands). These branches are merged and refined via a series of residual blocks with dense skip connections.}{The fusion model \( F_{\boldsymbol{\nu}}(\cdot, \cdot) \) is instantiated via a separate network for each resolution group (10\,m, 20\,m, and 60\,m), with approximately 200k parameters per model. }

\chg{The inputs to each branch are preprocessed with bilinear interpolation to restore native resolution, after applying blur and boxcar degradation in the training phase. At inference time, the RGB input is substituted with the 2.5 m super-resolved output, allowing the model to exploit high-frequency guidance it has never seen during training. The model is trained to minimize the \( \ell_1 \) reconstruction loss:}{The model is trained to minimize an \( \ell_1 \) reconstruction loss}
\[
\mathcal{L}_{\text{fusion}} = \left\| \widehat{\mathbf{X}}_{\text{LR}} - \mathbf{X}_{\text{LR}} \right\|_1,
\]
where \del{\chg{\( \mathbf{X}_{\text{HR}} \)}{\( \mathbf{X}_{\text{LR}} \)} is the native-resolution Sentinel-2 bands used as the supervision target, and} \del{\( \widehat{\mathbf{X}}_{\text{HR}} \)}
\add{
\begin{equation}
    \widehat{\mathbf{X}}_{\text{LR}} = \mathcal{F}_{\boldsymbol{\nu}}
    \left(
    \mathbf{X}_{\text{LR}}^\downarrow, \mathbf{x}_{\text{LR}}
    \right)
\end{equation}
}
is the predicted fused output. \del{This objective encourages the model to learn structure-preserving spectral fusion in a purely self-supervised fashion.}

\paragraph{\add{Detailed architecture }} 
\add{The details of architecure given in Fig.~\ref{fig:glpnn_pipeline}) are as follows:}

\add{{\textbf{(1) Convex mixer:}} The bands of the Sentinel-2 RGB, $\mathbf{x}_{LR}$, is combined into a  panchromatic proxy,}
\add{\[
\mathbf{p} = \sum_{c=1}^{3} w_c\, \mathbf{x}_{\mathrm{LR}}, 
\qquad w_c \ge 0,\ \sum_c w_c = 1,
\]
where \(\{w_c\}\) are softmax-constrained, learnable convex weights.}

\add{\textbf{(2) Sensor degradation model:} For each band \(i\), a Gaussian MTF blur \(\Phi_{\sigma_i}\) with reflection padding, followed by area downsampling and bicubic upsampling (\(\mathcal{D}_{r_i}, \mathcal{U}_{r_i}\)), produces the low-pass panchromatic image approximation}
\add{\[
\widetilde{\mathbf{p}}_i 
= \mathcal{U}_{r_i}\!\left(\mathcal{D}_{r_i}\left(\left(\Phi_{\sigma_i}(\mathbf{p}\right)\right)\right),
\]
and the high frequency details that need to be injected in the multi-spectral bands for each band is given as  \[\boldsymbol{\Delta}_i = \mathbf{p} - \widetilde{\mathbf{p}}_i.\]}

\add{
Here, the scaling factor \( r_i \) denotes the ratio between the native ground
sampling distance (GSD) of the \(i\)-th multispectral band and the panchromatic
proxy resolution. In practice, \(r_i\) determines the downsampling and
upsampling ratio applied in the simulated sensor degradation model
(Fig.~\ref{fig:glpnn_pipeline}). Specifically,
\[
r_i = \frac{\text{native GSD of band } i}{\text{target PAN GSD}}.
\]
For the Sentinel-2 configuration, we use three dedicated fusion branches
corresponding to the 10~m, 20~m, and 60~m resolution groups. We use \(r_i = 4, 8,\text{ and }24\)
for the 10 m, 20 m, and 60 m bands, respectively.
Each branch reconstructs its respective multispectral group to the 2.5~m
target resolution, guided by the super-resolved RGB proxy. 
}\add{Each blur kernel \(\Phi_{\sigma_i}\) is initialized from the nominal
MTF value of the corresponding Sentinel-2 band, converted to a Gaussian
standard deviation \(\sigma_i\) as a function of the Nyquist frequency and
the scale ratio \(r_i\).}

\add{\textbf{(3) DetailNet:} Each band’s detail is refined by a small residual network that operates on the concatenation of the GLP detail and the upsampled multispectral band group that is input to the fusion module, $\widetilde{\mathbf{x}}$ (Fig.~\ref{fig:glpnn_pipeline}):}
\add{\[
\widehat{\boldsymbol{\Delta}}_i 
= \boldsymbol{\Delta}_i + {\rm DetailNet}\!\left(\left[\boldsymbol{\Delta}_i,\ \widetilde{\mathbf{x}}\right]\right),
\]
where DetailNet consists of a \(3{\times}3\) convolution, LeakyReLU, two RRDB blocks, and a linear \(3{\times}3\) conv that outputs a one-channel residual map. The final layer has no activation to
retain the signed nature of the learned residual.}

\add{
\textbf{(4) GainNet:}
A per-band residual gain map is estimated from the concatenation of the
panchromatic proxy, its band-specific low-pass approximation, and the mean of
the upsampled multispectral inputs:
\[
g_i = 1 + {\rm GainNet}\!\left(
\left[\ \mathbf{p},\ 
\widetilde{\mathbf{p}}_i,\ 
\overline{\mathbf{x}}\ \right]
\right),
\]
where \(\overline{\mathbf{x}}\) denotes the per-pixel mean of the upsampled
multispectral group $\widetilde{\mathbf{x}}$. GainNet consists of a
shared two-layer convolutional subnetwork: a \(3{\times}3\) convolution with
LeakyReLU activation (\(0.2\) slope) followed by a \(1{\times}1\) convolution
that outputs a single-channel residual map \(\Delta g_i\). The final gain is
formulated as \(g_i = 1 + \Delta g_i\), providing an unbounded residual gain
around unity that scales the injected detail adaptively for each band.
}

\add{\textbf{(5) RefineNet and fusion:} The bandwise fusion applies the gain-modulated detail injection}
\add{\[
\mathbf{y}_i = \widetilde{\mathbf{x}}_i + g_i \cdot \widehat{\mathbf{\Delta}}_i ,
\]
where $\widetilde{\mathbf{x}}_i$ is the $i$th input multispectral band. The output \(\{\mathbf{y}_i\}\) is then stacked into into \(\mathbf{Y}\). A shallow refinement network then enforces inter-band consistency using the concatenation \([\mathbf{Y},\ \overline{\boldsymbol{\Delta}}_i]\), where \(\overline{\boldsymbol{\Delta}}_i\) is the mean refined across all detail bands $\{\boldsymbol{\Delta}_i\}$:}
\add{\[
\widehat{\mathbf{X}}_{\text{HR}} 
= \mathbf{Y} + {\rm RefineNet}\!\left(\left[\mathbf{Y},\ 
\overline{\boldsymbol{\Delta}}_i\right]\right),
\]
RefineNet consists of a \(3{\times}3\) convolution with LeakyReLU activation,
two residual blocks, and a final linear \(3{\times}3\) layer that outputs a
residual correction added back to \(\mathbf{Y}\).}

\add{\paragraph{Inference.} At inference, the fusion model receives inputs at their original resolutions (10 m, 20 m, or 60 m), but the native RGB input is replaced by a 2.5 m GSD super-resolved RGB image. This new input is low-pass filtered with the same per-band MTF kernels used in training to ensure frequency consistency. The network must generalize zero-shot to this enhanced spatial prior, as the super-resolved image is not used during the training phase~\cite {lanaras2018super}.}

\section{Experiments and Results}

\del{We will now evaluate the proposed pipeline (DiffFuSR) designed to super-resolve all 12 Sentinel-2 spectral bands to a 2.5-meter GSD. We first describe the datasets utilized for training the SR model and the fusion networks} \del{, followed by their respective training configurations.} \del{The performance of DiffFuSR is then benchmarked against established SR and pansharpening baselines. Evaluation is conducted through a comprehensive set of quantitative metrics assessing reflectance accuracy, spectral fidelity, spatial detail, and artifact generation, complemented by qualitative visual comparisons to demonstrate the efficacy of the super-resolved products.}

\subsection{Datasets}

To train and evaluate the proposed SR method, we use the following datasets:

\subsubsection{SEN2NAIP Dataset}
The SEN2NAIP dataset~\cite{aybar2024sen2naip} consists of two components: 
\begin{itemize}
    \item \textit{Cross-sensor dataset}: Paired imagery from Sentinel-2 and NAIP (National Agriculture Imagery Program), suitable for supervised learning tasks involving cross-sensor translation.
    \item \textit{Synthetic dataset}: Standalone NAIP images accompanied by transformation parameters for simulating Sentinel-2-like imagery. These parameters are based on four different histogram-based harmonization strategies.
\end{itemize}

\subsubsection{WorldStrat Dataset}
The WorldStrat dataset~\cite{cornebise2022open} contains HR RGB imagery acquired from the SPOT satellite. The data is globally distributed and tile-based, covering diverse geographic regions and land cover types. \del{It is designed to support machine learning tasks in remote sensing, such as SR and domain generalization.}

\subsubsection{Sentinel-2 Data}
The Sentinel-2 data used in this work consists of Level-2A products from the Sentinel-2 mission under the Copernicus program. These products provide 12 spectral bands at spatial resolutions of 10\,m, 20\,m, and 60\,m. All imagery has been curated to be cloud-free and atmospherically corrected, with spatial and temporal alignment across samples. \del{This dataset serves as the LR input for fusion and SR tasks.}

\subsection{\add{Super Resolution Model Training}\del{ Configurations}}

We train models using three primary data sources: airborne (NAIP), satellite-based (WorldStrat), and multispectral (Sentinel-2). These datasets are integrated into a unified diffusion-based SR pipeline.

\paragraph{Model Architecture}
The model architecture follows the design of \add{Wu et al.}~\cite{wu2023conditional} and comprises:
\begin{itemize}
  \item A degradation encoder for contrastive representation learning. \add{(2.6M parameters)}
  \item A spatial encoder-decoder that is implemented using an RRDB-Net encoder with 23 residual blocks and 64 feature channels.\add{(16.7M parameters)}
  \item A diffusion UNet with 5 residual blocks and channel multipliers \{1, 2, 2, 4, 8\}. \add{(148.3M parameters)}
  \item Diffusion configuration: we use 1000 diffusion steps, with a cosine noise schedule \cite{wu2023conditional}. The model predicts the original clean image $\mathbf{x}_0$ directly.
  \item Exponential moving average of weights with a decay rate of 0.999.
  
\end{itemize}

\paragraph{Training Configuration}
All models were trained for 1000 epochs using the Adam optimizer ($\beta_1=0.9$, $\beta_2=0.999$) with an initial learning rate of  $1\times 10^{-4}$ , decayed by a factor of 0.5 at epochs 200, 400, 600, and 800. We used a batch size of 8 and image patch size of 256$\times$256. Data augmentation included random cropping and flipping.

\paragraph{NAIP (Harmonized and Unharmonized)}
Training used samples from the synthetic set of SEN2NAIP~\cite{aybar2024sen2naip}. In the harmonized variant, on-the-fly histogram harmonization was applied during data loading. For unharmonized training, this step was omitted. Both configurations used four-channel multispectral inputs. Degradation modeling applied anisotropic Gaussian blur (kernel size 21, $\sigma \in [2.0, 4.0]$) and additive Gaussian noise with standard deviation 25.

\paragraph{WorldStrat}
A separate model was trained on globally distributed tiles from the WorldStrat dataset~\cite{cornebise2022open}. As with NAIP, four-channel input was used and the same model architecture and optimizer configuration were applied. \del{This setup facilitates assessment of generalization across diverse geographies and domains.}

\subsection{Fusion Model Training}

\del{We train a cross-sensor fusion model to enhance the spatial resolution in Sentinel-2 multispectral imagery by leveraging HR inputs and multi-resolution contextual information. The fusion model operates over three spatial scales corresponding to Sentinel-2 band groups: 10\,m, 20\,m, and 60\,m resolutions.}

\paragraph{Model Architecture}

\chg{The model is a multi-branch variant of RRDB-Net\mbox{~\cite{wang2018esrgan}}, consisting of three dedicated fusion modules:}%
{The proposed fusion model is implemented as three dedicated MTFNetFusion modules, one per resolution group (10\,m, 20\,m, 60\,m):}

\begin{itemize}
\item A \textit{10\,m branch} that fuses RGB and near infrared bands (bands 2, 3 4 and 8). \add{(202k parameters)}
\item A \textit{20\,m branch} that fuses the red edge and short wave infrared bands (bands 5, 6, 7, 8A, 11, and 12). \add{(202k parameters)}
\item A \textit{60\,m branch} that fuses coastal aerosol and water vapor bands (bands 9 and 10). \add{(202k parameters)}
\end{itemize}

\chg{Each branch uses an RRDB-Net with 5 residual blocks, 64 base channels, and group convolutions of 32.}{Each branch follows a MTF-GLP design, as specified in Sec. \ref{sec:fusion}.}
\del{The fusion signal (i.e., the super-resolved S2 RGB bands) is concatenated with interpolated LR multispectral inputs and used to reconstruct HR outputs at each scale.}

\paragraph{Training Details}

Three models are trained using separate optimizers for each spatial branch (10\,m, 20\,m, 60\,m), each using Adam with a learning rate of $1\times10^{-4}$. \del{A standard L1 loss is applied independently at each scale.}
\del{The models are optimized with an $L_1$ loss:}
\del{\[
\mathcal{L}_{\text{fusion}} = \big\| \widehat{\mathbf{X}}_{\text{HR}} - \mathbf{X}_{\text{HR}} \big\|_1.\]}
\del{where $\mathbf{X}_{\text{HR}}$ is the native-resolution reference.}

\paragraph{Validation Metric}

\chg{During validation, the model's outputs are compared to ground truth using the ERGAS metric, computed at each resolution.}{Validation is performed using ERGAS at each resolution.}
\chg{Additionally, a Gram-Schmidt pan-sharpening baseline is computed for all outputs.}{Additionally, multiple pan-sharpening baselines are computed for comparison.}

\add{\paragraph{Inference Time}
The inference time for the diffusion model is 1.74 seconds, and for the fusion model is  0.0161 seconds.
}
\subsection{Baseline Methods}

Existing state-of-the-art methods do not directly address the task of super-resolving all Sentinel-2 bands to a unified resolution. Since DiffFuSR is modular, we evaluate its components in two stages. First, we compare its RGB SR module against established RGB-specific SR baselines. Second, we evaluate the fusion module separately by comparing it to a suitable multispectral fusion baseline that combines super-resolved RGB with the remaining bands. \del{This decomposition allows for a systematic assessment of each component relative to current state-of-the-art approaches.}

\subsubsection{Super resolution baselines}

\del{We evaluate our}\add{The} proposed SR approach \add{is evaluated} against a diverse set of existing models.\del{, representing a spectrum of training philosophies.} The \textit{OpenSR baseline}~\cite{aybar2024sen2naip} serves as a strong reference point, particularly due to its use of degradation-based training on NAIP data. The \textit{LDM\_Baseline}~\cite{10887321} employs a latent diffusion model, applying generative priors to RGB upsampling\del{, but suffers from instability and hallucination in complex scenes}. \textit{Superimage} adopts a pixel-wise regression framework optimized for spectral fidelity. 
\del{\textit{Satlas}\mbox{~\cite{wolters2023zooming}}, trained using a generative adversarial network\mbox{~\cite{wang2018esrgan}} on cross-domain imagery, tends to produce sharper images but depicts poor spectral integrity.}
\add{We compare against \textit{Satlas}~\cite{wolters2023zooming}, a GAN-based model trained on cross-domain imagery.}
Additionally, we compare with \textit{SR4RS} \cite{cresson2022sr4rs}, a regression-based model targeting RGB SR via auxiliary feature learning\del{, but which introduces spatial inconsistencies in challenging regions}.

\subsubsection{Fusion model baseline}

\del{To benchmark the effectiveness of the learned fusion network, we implement a classical Gram-Schmidt (GS) pansharpening baseline\mbox{~\cite{loncan2015hyperspectral,vivone2014critical}}. This method fuses LR multispectral bands with a HR RGB image, simulating a panchromatic band by averaging the RGB channels. The GS transformation aligns the spectral content of the MS input with the spatial detail of the pan proxy, producing a HR output without learning.}

\del{For fair comparison, we apply GS using the same degraded multispectral bands and super-resolved RGB image used by our model at inference time. The GS baseline thus serves as a non-parametric alternative guided by the same spatial prior, highlighting the value added by learned fusion.}

\add{To benchmark the effectiveness of the proposed fusion network (GLP-NN), we implement Gram-Schmidt (GS) pansharpening ~\cite{vivone2014critical}, the intensity-hue-saturation (IHS) transform, principal component analysis (PCA), and the classical GLP  with MTF filters and full-scale regression injection
model (GLP) method ~\cite{vivone2018full}. Furthermore, following~\cite{deng2022machine}, we include their state-of-the-art FusionNet as a learning-based baseline. Although FusionNet was originally designed for generic pansharpening, we retrain it under the same conditions as our proposed fusion module to ensure a fair comparison.}

\subsection{Metrics}
\subsubsection{Super Resolution}

We evaluate the proposed SR model using the OpenSR-test benchmark~\cite{opensrtest}, which offers cross-sensor datasets (NAIP, SPOT, Ven$\mu$S, SPAIN CROPS, SPAIN URBAN) with accurately aligned Sentinel-2 LR and HR reference imagery. The benchmark assesses performance through a set of intuitive metrics, grouped into three categories:

\emph{Consistency} evaluates whether the super-resolved image, when downsampled, preserves key properties of the original Sentinel-2 input. It includes: (i) \textit{reflectance}, measuring radiometric similarity via chosen distance metric; (ii) \textit{spectral}, quantifying preservation of spectral signatures through spectral angle distance (SAD); and (iii) \textit{spatial}, assessing alignment using phase correlation.

\emph{Synthesis} captures how much meaningful high-frequency detail the model introduces beyond simple interpolation. \del{Higher synthesis values indicate better recovery of fine structures such as edges, textures, and small objects that are absent in the low-resolution input.}

\emph{Correctness} analyzes whether the generated details are realistic. It includes: (i) \textit{hallucinations}, quantifying spurious details not present in the true HR reference; (ii) \textit{omissions}, measuring genuine HR details that the model failed to recover; and (iii) \textit{improvements}, reflecting correctly added fine-scale structures that were not visible in the low-resolution input. \del{Ideally, hallucination and omission scores should be low, while improvement should be high.}

The distance metric chosen for calculation can be either normalized difference distance, LPIPS \cite{zhang2018unreasonable}, or CLIP distance \cite{radford2021learning}. \del{Using all three gives us physical, perceptual, and semantic validation, respectively, which makes OpenSR-test uniquely strong compared to typical SR benchmarks.} For full mathematical definitions and implementation details, we refer to the OpenSR-test paper~\cite{opensrtest}.

\subsubsection{Fusion Pipeline}
To evaluate the quality of the SR fusion outputs in the absence of ground truth \del{(GT)} HR Sentinel-2 imagery, we adopt a no-reference validation strategy. Specifically, we downsample the SR outputs back to the original Sentinel-2 resolution and perform comparisons against the native Sentinel-2 inputs. \del{This indirect supervision enables us to assess fidelity and consistency across multiple quality metrics.}
\ifshowchanges

\del{We conduct a quantitative comparison of different fusion methods, neural network \del{(NN)}versus GS-based fusion across the 12 Sentinel-2 bands.} \del{Evaluations are carried out using the following metrics:}
\begin{itemize}
    \item \del{\textit{R\textsuperscript{2}}}: \del{Coefficient of determination, indicating how well predicted reflectance values approximate the actual values. Higher R\textsuperscript{2}-values is better, indicating a strong alignment between the predicted and actual values.}
    \item \del{\textit{Cross-Correlation}}: \del{Measures the linear relationship between the predicted and original signal. Values close to one indicate strong similarity.}
    \item \del{\textit{SSIM (Structural Similarity Index)}}: \del{Assesses perceptual similarity in terms of structure and luminance. Higher values denote more faithful reconstructions \mbox{\cite{blau2018perception}}.}
    \item \del{\textit{PSNR (Peak Signal-to-Noise Ratio)}}: \del{Evaluates the ratio between the maximum possible signal and the power of corrupting noise. Higher PSNR suggests better fidelity  \mbox{\cite{blau2018perception}}.}
    \item \del{\textit{MSE (Mean Squared Error)}}: \del{Captures the average squared difference between the downsampled SR image and the original Sentinel-2 image. Lower values are preferred.}
    \item \del{\textit{ERGAS (Relative Dimensionless Global Error in Synthesis)}}: \del{A global error metric specific to remote sensing fusion tasks. Lower values indicate better spectral and spatial consistency  \mbox{\cite{wald2002data}}.}
\end{itemize}
\else
\fi
\add{Evaluations are carried out using the following metrics: coefficient of determination (\textit{R\textsuperscript{2}}), cross-correlation, Structural Similarity Index (SSIM) \cite{blau2018perception}, Peak Signal-to-Noise Ratio (PSNR) \cite{blau2018perception}, Mean Squared Error (MSE), and ERGAS (Relative Dimensionless Global Error in Synthesis) \cite{wald2002data}.}


\subsection{Qualitative Evaluation}

\begin{figure*}[htbp]
    \centering

    \vspace{0.5em}


    \includegraphics[width=0.80\textwidth, trim=0 328 0 0, clip]{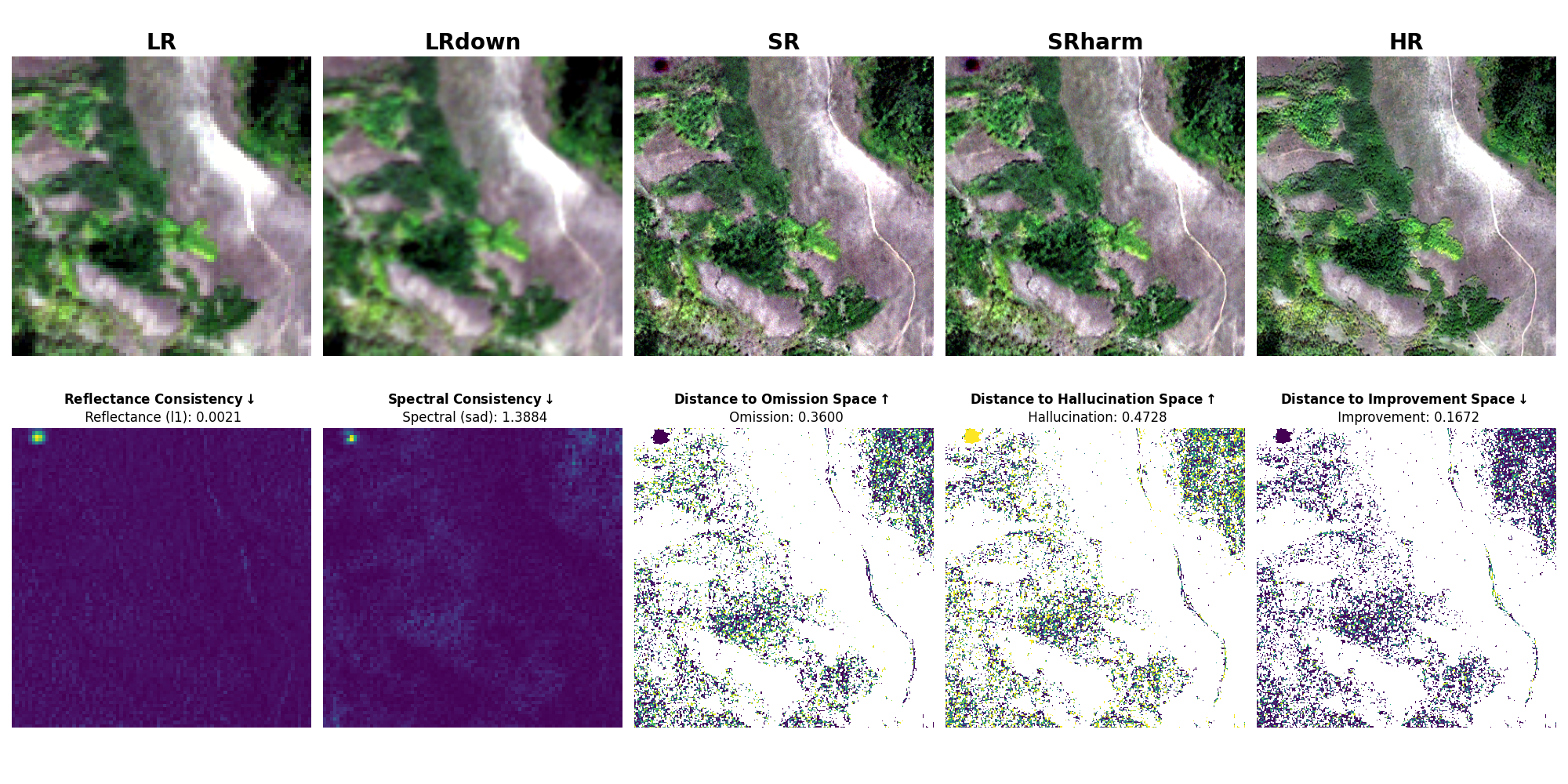}
 
    \vspace{0.5em}


    \includegraphics[width=0.80\textwidth, trim=0 328 0 0, clip]{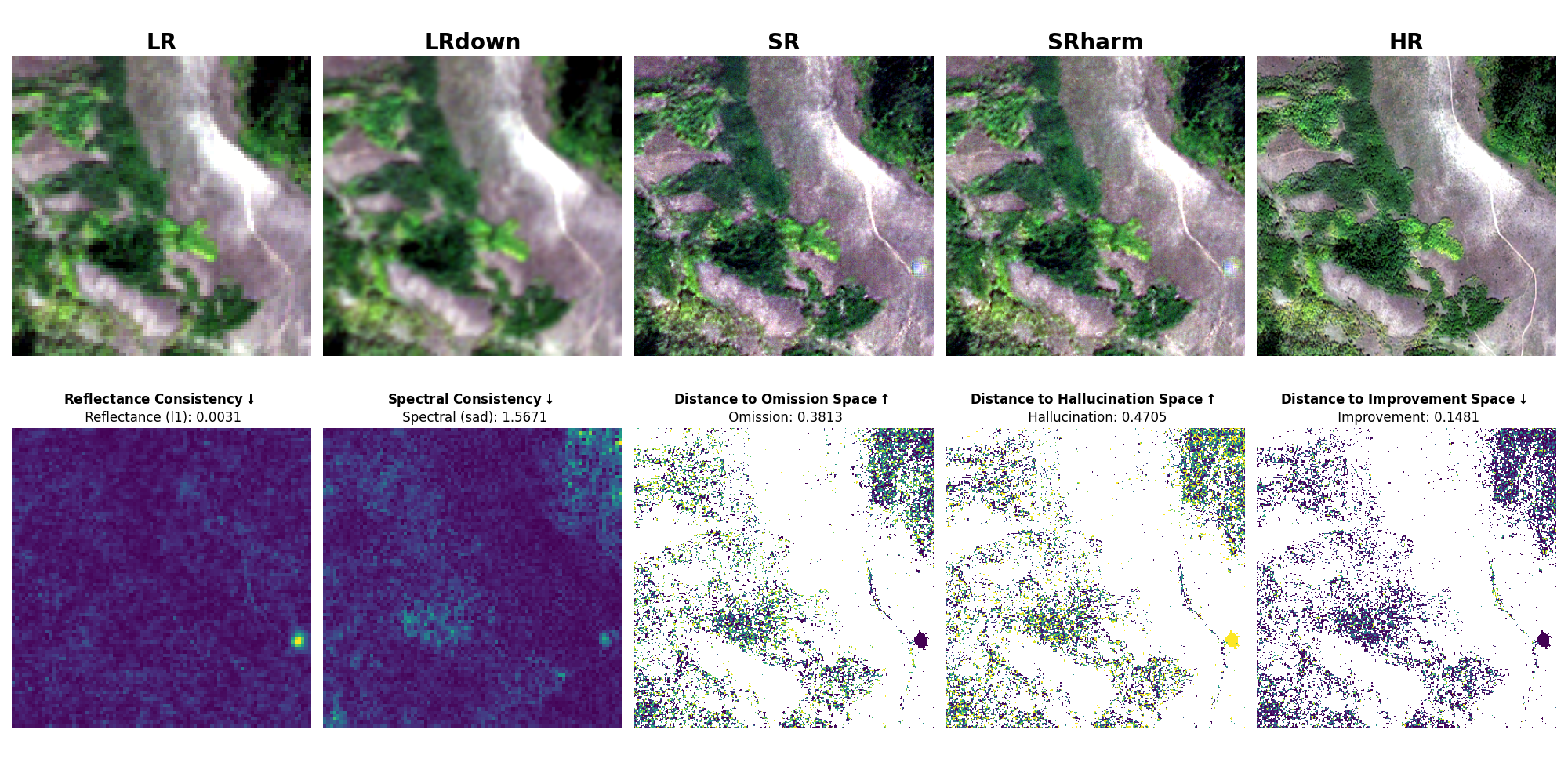}
 
    \vspace{0.5em}


    \includegraphics[width=0.80\textwidth, trim=0 328 0 0, clip]{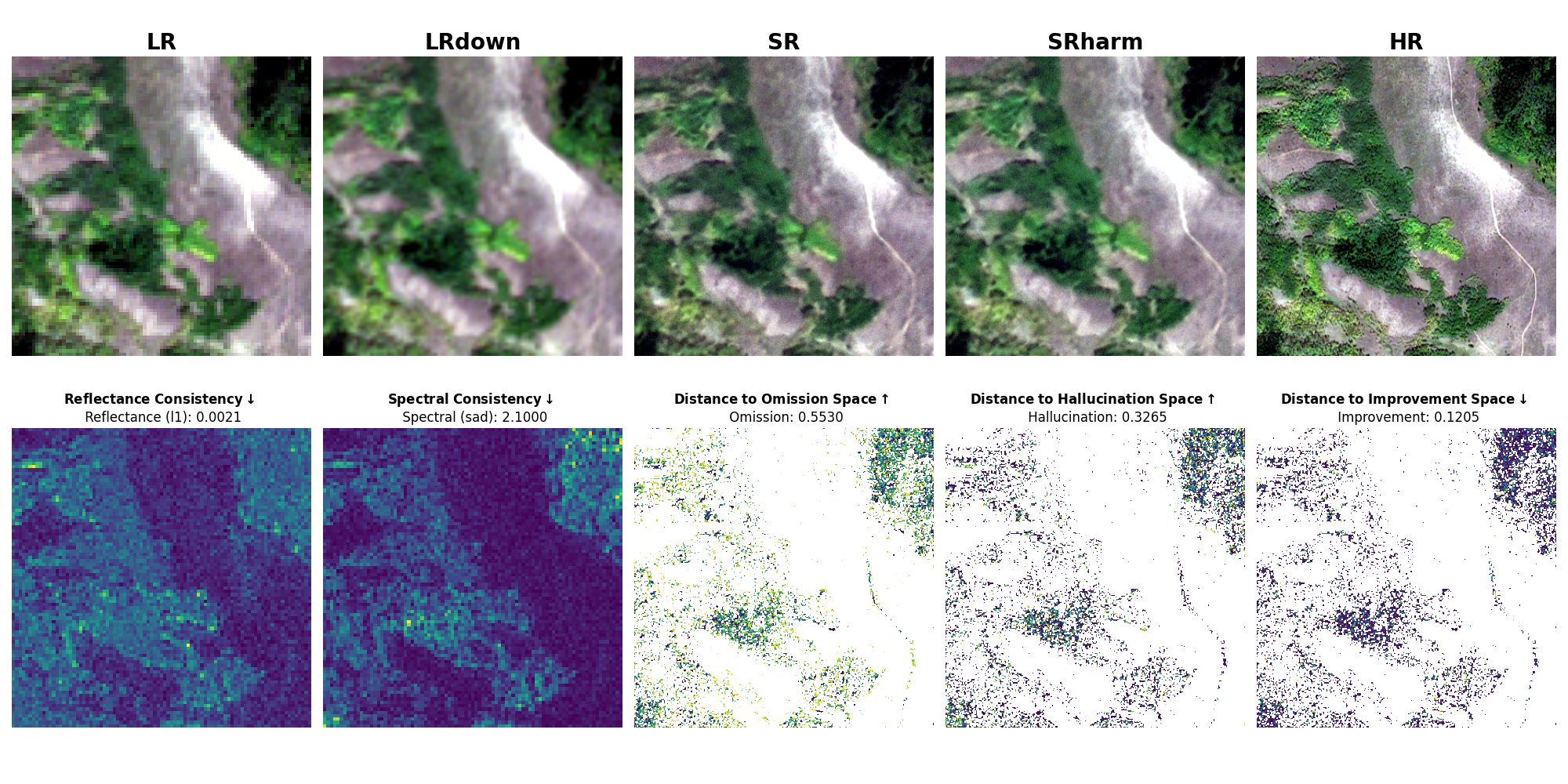}
 
\caption{Qualitative comparison of RGB SR results (vegetated scene). Rows correspond to models trained on harmonized NAIP, unharmonized NAIP, and WorldStrat. Top sub-row shows image outputs: LR input, bicubic downsampled variant, SR result, harmonized SR, and ground truth. Bottom sub-row shows reflectance consistency (L1), spectral consistency (SAD), and distances to omission, hallucination, and improvement spaces. The harmonized NAIP model yields the best visual and quantitative performance in this vegetated context.}
    \label{fig:rgbSR1}
\end{figure*}

\begin{figure*}[htbp]
    \centering

    \includegraphics[width=0.80\textwidth, trim=0 328 0 0, clip]{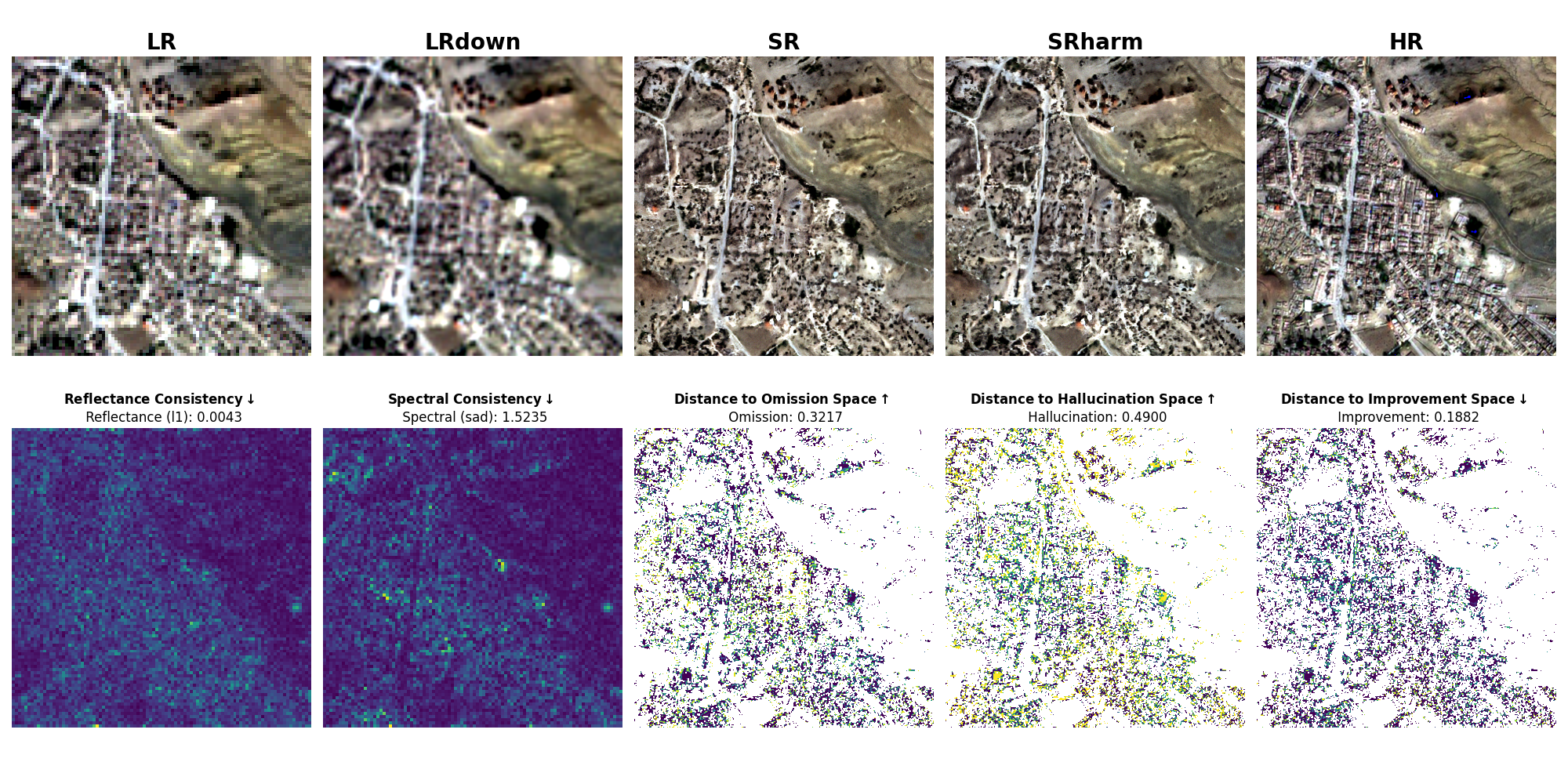}
 

    \includegraphics[width=0.80\textwidth, trim=0 328 0 0, clip]{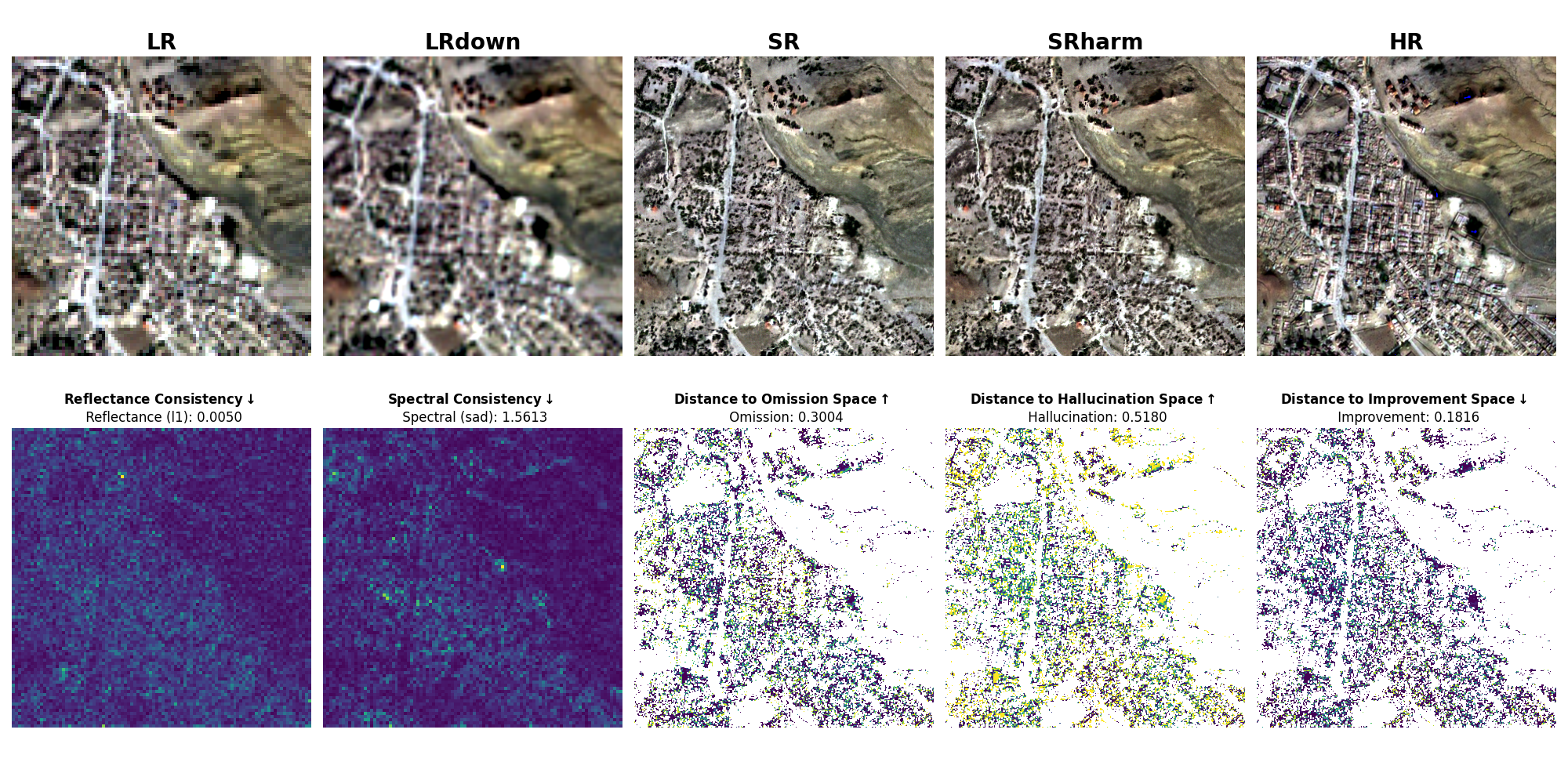}
 

    \includegraphics[width=0.80\textwidth, trim=0 328 0 0, clip]{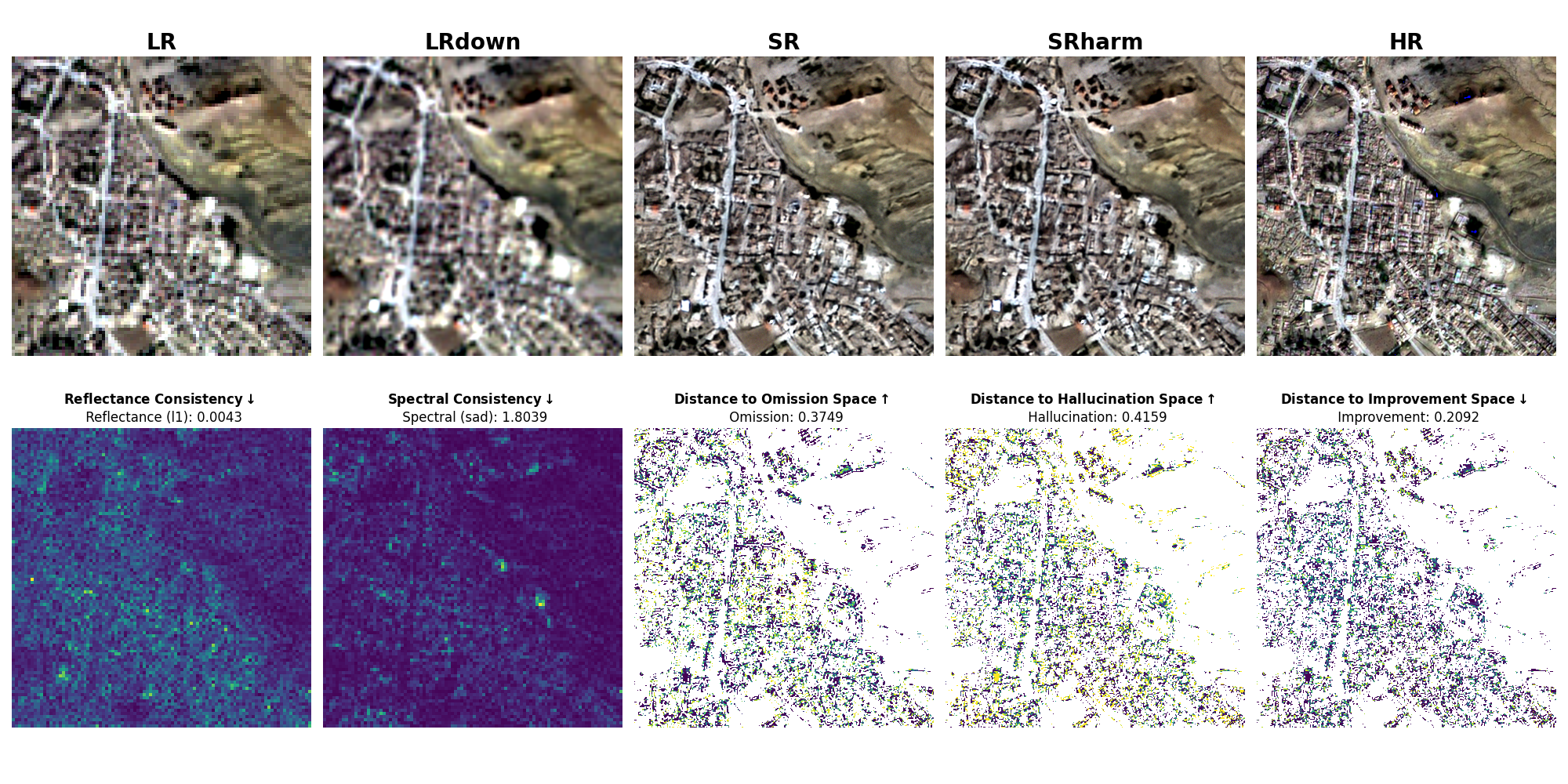}
 
    \caption{Qualitative comparison of RGB SR results (urban scene). As in Figure~\ref{fig:rgbSR1}, rows represent different training sources. In contrast to the vegetated example, the WorldStrat-trained model produces sharper and more realistic structures in buildings and roads, suggesting stronger generalization in urban domains despite slightly higher spectral error.}
    \label{fig:rgbSR2}
\end{figure*}

\subsubsection{Super resolution}
Figs.~\ref{fig:rgbSR1} and~\ref{fig:rgbSR2} illustrate qualitative comparisons of the proposed SR model trained on three different datasets: harmonized NAIP, unharmonized NAIP, and WorldStrat. Each figure is organized into three major rows (one per dataset), where \add{each}\del{the top} row shows \del{reconstructed} images, including the LR input, downsampled version, SR output, harmonized SR output, and the HR reference. \del{The second row visualizes per-pixel evaluation maps for reflectance consistency, spectral consistency (SAD), omission space, hallucination space, and improvement space, offering visual insight into the model’s behavior.}

\begin{table*}[t]
\centering
\caption{Normalized difference–based distance calculation (\textbf{Bold} = best, \underline{Underlined} = second best; ↓ lower is better, ↑ higher is better). “Avg.\ Rank” summarizes each model’s mean rank across all seven metrics.}
\label{tab:nd}
\scalebox{0.82}{
\begin{tabular}{lccccccccc}
\toprule
\textbf{Model} 
  & Reflectance $\downarrow$ 
  & Spectral $\downarrow$ 
  & Spatial $\downarrow$ 
  & Synthesis $\uparrow$ 
  & Hallucination $\downarrow$ 
  & Omission $\downarrow$ 
  & Improvement $\uparrow$ 
  & Avg.\ Rank \\
\midrule
ldm\_baseline          
  & 0.0505 ± 0.0161 
  &  9.6923 ± 2.1742 
  & 0.0715 ± 0.0679 
  & \best{0.0285 ± 0.0307} 
  & 0.6067 ± 0.2172 
  & 0.3088 ± 0.1786 
  & 0.0845 ± 0.0428 
  & 5.57 \\
opensrmodel            
  & 0.0031 ± 0.0018 
  &  1.2632 ± 0.5878 
  & 0.0114 ± 0.0111 
  & 0.0068 ± 0.0044 
  & \second{0.3431 ± 0.0738} 
  & 0.4593 ± 0.0781 
  & \best{0.1976 ± 0.0328} 
  & 4.00 \\
satlas                 
  & 0.0489 ± 0.0086 
  & 12.1231 ± 3.1529 
  & 0.2742 ± 0.0748 
  & \second{0.0227 ± 0.0107} 
  & 0.8004 ± 0.0641 
  & \best{0.1073 ± 0.0393} 
  & 0.0923 ± 0.0266 
  & 5.71 \\
sr4rs                  
  & 0.0396 ± 0.0198 
  &  3.4044 ± 1.6882 
  & 1.0037 ± 0.1520 
  & 0.0177 ± 0.0083 
  & 0.7274 ± 0.0840 
  & \second{0.1637 ± 0.0572} 
  & 0.1089 ± 0.0292 
  & 5.42 \\
superimage             
  & 0.0029 ± 0.0009 
  &  1.5672 ± 1.0692 
  & 0.0132 ± 0.1131 
  & 0.0046 ± 0.0027 
  & \best{0.2026 ± 0.0692} 
  & 0.6288 ± 0.0754 
  & 0.1686 ± 0.0302 
  & 5.00 \\
DiffFuSR (NAIP Harm)   
  & \best{0.0024 ± 0.0009} 
  & \best{1.1103 ± 0.6197} 
  & \second{0.0052 ± 0.0092} 
  & 0.0083 ± 0.0043 
  & 0.4580 ± 0.1020 
  & 0.3543 ± 0.0836 
  & \second{0.1876 ± 0.0344} 
  & \best{2.85} \\
DiffFuSR (NAIP Unharm) 
  & 0.0038 ± 0.0008 
  & \second{1.2320 ± 0.7256} 
  & 0.0053 ± 0.0093 
  & 0.0086 ± 0.0042 
  & 0.4927 ± 0.1095 
  & 0.3373 ± 0.0870 
  & 0.1701 ± 0.0351 
  & 3.857 \\
DiffFuSR (Worldstrat)  
  & \second{0.0026 ± 0.0009} 
  &  1.3277 ± 0.7205 
  & \best{0.0032 ± 0.0074} 
  & 0.0072 ± 0.0037 
  & 0.3915 ± 0.1209 
  & 0.4248 ± 0.1058 
  & 0.1837 ± 0.0409 
  & \second{3.57} \\

\bottomrule
\end{tabular}
}
\end{table*}

\begin{table*}[t]
\centering
\caption{CLIP Distance (\textbf{Bold} = best, \underline{Underlined} = second best; ↓ lower is better, ↑ higher is better). “Avg.\ Rank” summarizes each model’s mean rank across all seven metrics.}
\label{tab:clip}
\scalebox{0.82}{
\begin{tabular}{lcccccccc}
\toprule
\textbf{Model} 
  & Reflectance $\downarrow$ 
  & Spectral $\downarrow$ 
  & Spatial $\downarrow$ 
  & Synthesis $\uparrow$ 
  & Hallucination $\downarrow$ 
  & Omission $\downarrow$ 
  & Improvement $\uparrow$ 
  & Avg.\ Rank \\
\midrule
ldm\_baseline          
  & 0.1239 ± 0.0405 
  & 12.8441 ± 2.7508 
  & 0.0717 ± 0.0683 
  & 0.0409 ± 0.0290 
  & 0.5963 ± 0.3055 
  & \best{0.2327 ± 0.2238} 
  & 0.1710 ± 0.1435 
  & 5.57 \\
opensrmodel            
  & 0.0076 ± 0.0046 
  &  1.9739 ± 1.0507 
  & 0.0118 ± 0.0108 
  & 0.0194 ± 0.0120 
  & \second{0.1052 ± 0.0590} 
  & 0.6927 ± 0.1343 
  & 0.2021 ± 0.0854 
  & 5.00\\
satlas                 
  & 0.1197 ± 0.0233 
  & 15.1521 ± 2.9876 
  & 0.2766 ± 0.0741 
  & \best{0.0648 ± 0.0302} 
  & 0.6996 ± 0.2058 
  & \best{0.0872 ± 0.0947} 
  & 0.2132 ± 0.1393 
  & 5.14 \\
sr4rs                  
  & 0.0979 ± 0.0509 
  & 22.4905 ± 2.1168 
  & 1.0099 ± 0.0439 
  & \second{0.0509 ± 0.0237} 
  & 0.3099 ± 0.1704 
  & 0.3486 ± 0.1753 
  & 0.3415 ± 0.1042 
  & 5.28 \\
superimage             
  & \best{0.0068 ± 0.0016} 
  &  1.8977 ± 1.1053 
  & \best{0.0004 ± 0.0032} 
  & 0.0130 ± 0.0073 
  & \best{0.0610 ± 0.0305} 
  & 0.8524 ± 0.0586 
  & 0.0866 ± 0.0395 
  & 4.28 \\
DiffFuSR (NAIP Harm)   
  & \second{0.0069 ± 0.0032} 
  & \best{1.6739 ± 0.8602} 
  & 0.0061 ± 0.0098 
  & 0.0238 ± 0.0120 
  & 0.2883 ± 0.2333 
  & 0.3358 ± 0.2214 
  & \second{0.3759 ± 0.1444} 
  & \best{3.00} \\
DiffFuSR (NAIP Unharm) 
  & 0.0109 ± 0.0030 
  &  2.0734 ± 1.0205 
  & 0.0061 ± 0.0096 
  & 0.0245 ± 0.0118 
  & 0.3272 ± 0.2400 
  & 0.2929 ± 0.2169 
  & \best{0.3799 ± 0.1475} 
  & 3.92 \\
DiffFuSR (Worldstrat)  
  & 0.0075 ± 0.0031 
  & \second{1.7327 ± 0.8545} 
  & \second{0.0042 ± 0.0086} 
  & 0.0205 ± 0.0105 
  & 0.2612 ± 0.2544 
  & 0.4646 ± 0.2773 
  & 0.2742 ± 0.1426 
  & \second{3.71} \\

\bottomrule
\end{tabular}
}
\end{table*}

In Fig.~\ref{fig:rgbSR1}, which features a forested scene with green tree cover and bare land, the harmonized NAIP model clearly yields the sharpest and most visually realistic reconstruction. Both the SR image and the harmonized version closely resemble the HR reference, with fine details preserved. This is supported by the quantitative \del{overlays}\add{values}, where reflectance and spectral consistency metrics are lowest (L1: 0.0021, SAD: 1.3884) and omission and hallucination errors are moderate. The unharmonized NAIP variant performs reasonably well but introduces more spectral error and slightly higher hallucination (SAD: 1.5671, HA: 0.4705). The WorldStrat-trained model appears perceptually blurrier and less structurally aligned, despite having similar reflectance L1, and its SAD score (2.10) and omission space distance (0.5530) indicate weaker spectral fidelity. \del{This case highlights the strength of NAIP-derived training, especially in harmonizing the reconstruction of vegetated regions.} The likely explanation is that there is a heavy focus on areas with vegetation in NAIP dataset, whereas the WorldStrat dataset has a slight urban area tilt.

Conversely, Fig.~\ref{fig:rgbSR2} presents an urban scene with buildings, where the performance trend shifts. Here, the WorldStrat-trained model produces the most visually convincing output with edges being sharp, and structural details like roof boundaries and roads being clearly resolved. Despite having a slightly higher spectral error (SAD: 1.8039), its visual quality surpasses that of both NAIP-trained models. The harmonized NAIP model underperforms in this case, with noticeable blurring and misalignment around fine structures, and a higher hallucination score (0.4900). \del{This result demonstrates that while NAIP performs well in vegetated regions, it may struggle in built-up areas where WorldStrat’s diverse urban coverage offers an advantage.}

\add{The SR images in Fig.~\ref{fig:rgbSR1} and Fig.~\ref{fig:rgbSR2} are histogram-matched with the HR ground truth image to obtain SRharm which is the harmonized
SR output. This additional SR output is used as a diagnostic tool: a minimal visual change
post-matching indicates high intrinsic spectral fidelity.}

\subsubsection{Fusion module}
\begin{figure}[htbp]
    \centering

    \subfigure{\includegraphics[width=0.151\textwidth]{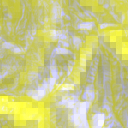}}
    \hfill
    \subfigure{\includegraphics[width=0.151\textwidth]{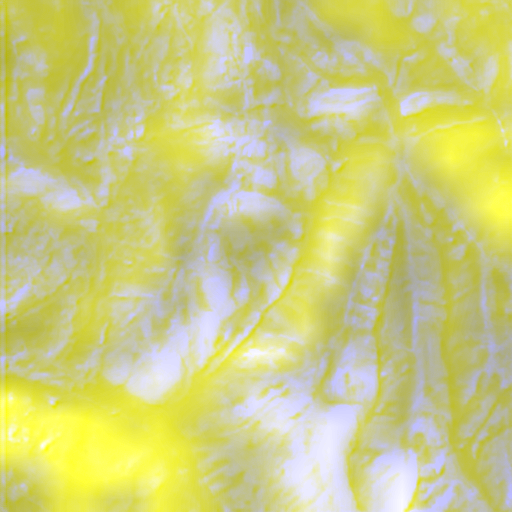}}
    \hfill
    \subfigure{\includegraphics[width=0.151\textwidth]{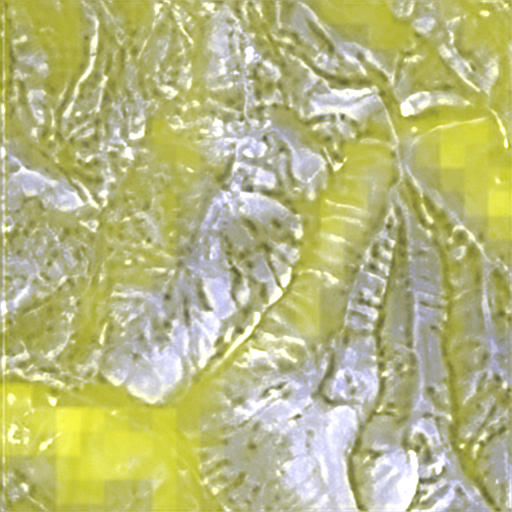}}
    \begin{tabular}{ccc}
        \makebox[0.15\textwidth]{\small{Low Resolution}} &
        \makebox[0.15\textwidth]{\small{GLP \cite{vivone2018full}}} &
        \makebox[0.15\textwidth]{\small{GLP NN (ours)}}
    \end{tabular}

    \subfigure{\includegraphics[width=0.151\textwidth]{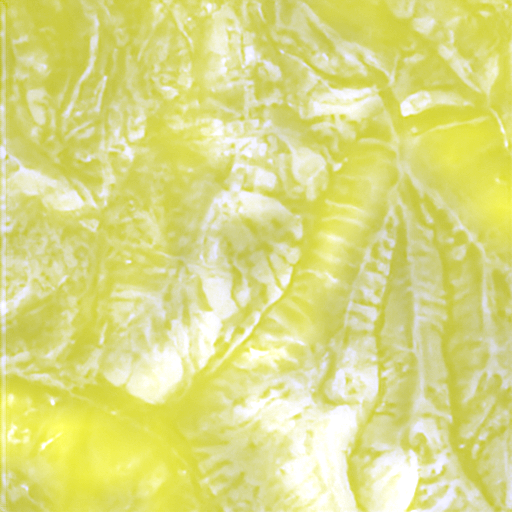}}
    \hfill
    \subfigure{\includegraphics[width=0.151\textwidth]{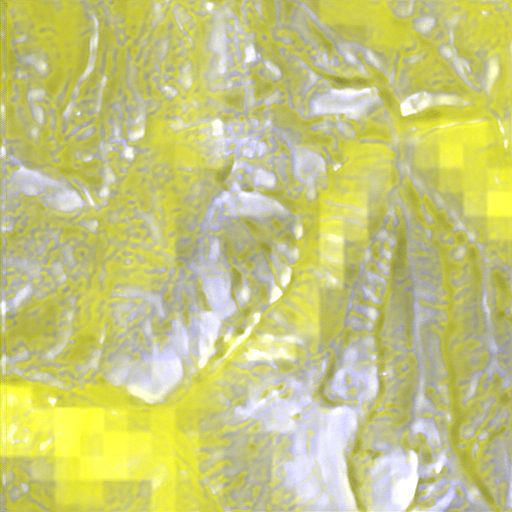}}
    \hfill
    \subfigure{\includegraphics[width=0.151\textwidth]{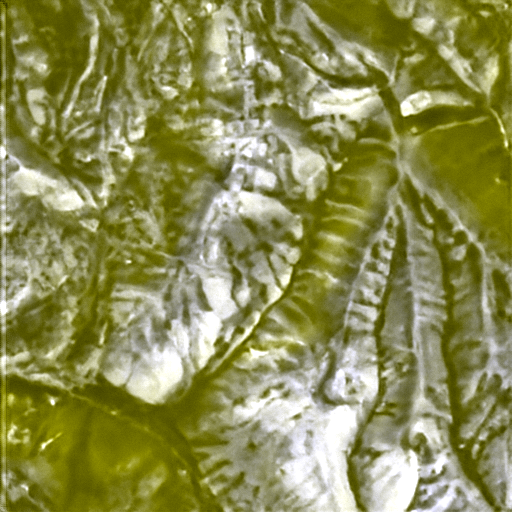}}
    \begin{tabular}{ccc}
        \makebox[0.15\textwidth]{\small{IHS}} &
        \makebox[0.15\textwidth]{\small{Fusion Network \cite{deng2022machine}}} &
        \makebox[0.15\textwidth]{\small{Gram Schmidt \cite{dalla2015global}}}
    \end{tabular}

    \caption{\add{Fusion output comparisons across 60\,m atmospheric bands. The figure show the Low Resolution input, Generalized laplacian pyramid (GLP) baseline method \cite{vivone2018full}, our GLP inspired neural network, intensity-hue-saturation (IHS) transform, and Fusion network baseline \cite{deng2022machine} and  Gram Schmidt \cite{dalla2015global} method.}}

    \label{fig:fusion}
\end{figure}

Fig.~\ref{fig:fusion} evaluates the proposed learnable fusion model. Here, we only use the harmonised NAIP-based SR model to create the RGB super-resolved image that is further used for fusion. \del{The images represent the LR input, GS fusion, and our neural network-based fusion, applied to 60 m bands and visualized in natural and false color. Compared to GS, which often distorts the structure and exaggerates spectral contrast, the fusion model output is visually more consistent with the low-resolution signal while enhancing spatial detail. Particularly in the 60 m bands, the fusion model preserves smooth gradients and subtle features that are entirely lost in GS. These results confirm that the learned fusion pipeline generalizes well and substantially outperforms classical pansharpening.} \add{ The columns represent the Low Resolution input, Generalized laplacian pyramid (GLP) baseline method \cite{vivone2018full}, our GLP inspired neural network, intensity-hue-saturation (IHS) transform, and Fusion network baseline \cite{deng2022machine} and Gram Schmidt \cite{dalla2015global} method applied to 60 m bands and visualized in RGB, where the red and green channels use the Water Vapour band (B10) and the blue channel uses the Coastal Aerosol band (B1). Compared to the baselines, which often distort the structure and exaggerate spectral contrast, our GLP inspired fusion networks output is visually more consistent with the low-resolution signal while enhancing spatial detail.}

\del{Together, these examples highlight the importance of dataset diversity and the advantage of learning-based fusion. NAIP excels in natural environments, WorldStrat performs better in urban settings, and neural fusion consistently surpasses traditional methods.}

\subsection{Quantitative Evaluation}
%
\del{To assess the efficacy of the proposed SR pipeline, we conducted a comprehensive quantitative evaluation across three distance metrics: Normalized Difference, CLIP, and LPIPS\mbox{~\cite{opensrtest}}. These metrics collectively assess reflectance preservation, spectral and spatial fidelity, and perceptual realism. In addition, we summarize the performance of each method using average rank.} To obtain the average rank, each model is ranked individually for every metric: for metrics where lower values indicate better performance (Reflectance, Spectral, Spatial, Hallucination, and Omission), ranks are assigned from lowest (rank 1) to highest (rank 8); for metrics where higher values are better (Synthesis and Improvement), ranks are assigned from highest (rank 1) to lowest (rank 8). The average rank for each model is then computed as the arithmetic mean of its ranks across all seven metrics, providing a single aggregated measure of overall performance.

\subsubsection{Performance on Normalized Difference Metrics}

The results in \del{Table~\ref{tab:nd}} show that the proposed diffusion model, trained on harmonized NAIP data, achieves best-in-class performance in multiple key metrics, including average rank \add{(Table~\ref{tab:nd})}. Specifically, it achieves the lowest reflectance error ($0.0024$), highest spectral fidelity ($1.1103$), and the second-best spatial alignment ($0.0052$). Its synthesis score ($0.0083$) and omission ($0.3543$) are high among real competitors, while hallucination errors remain moderate compared to baselines. \del{Notably, our model’s improvement score ($0.1876$) confirms its strong ability to enrich LR inputs with high-frequency, realistic details.}

\subsubsection{CLIP Distance Analysis}

Table~\ref{tab:clip} further confirms the robustness of DiffFuSR trained with NAIP harmonized data. Reflectance performance is essentially tied with Superimage, and our model delivers the best spectral score ($1.6739$). Its spatial error ($0.0061$) is among the lowest, while the improvement metric peaks at $0.3759$, indicating an enhanced ability to reconstruct semantically meaningful content.

\subsubsection{Insights from LPIPS Metric}

\del{As illustrated in Table~\ref{tab:lpips}, }LPIPS-based analysis shows that the proposed method not only achieves strong perceptual fidelity but also minimizes over-synthesis \add{(Table~\ref{tab:lpips})}. It delivers a moderate hallucination score ($0.1149$), confirming that our model avoids generating too many spurious details. Additionally, it achieves the best improvement score ($0.5383$), reaffirming its effectiveness in reconstructing rich, high-frequency features from low-resolution imagery.

\del{It is interesting to note that the DiffFuSR-based models achieve the best average ranking across the three Tables.}

\begin{table*}[t]
\centering
\caption{LPIPS Distance (\textbf{Bold} = best, \underline{Underlined} = second best; ↓ lower is better, ↑ higher is better). “Avg.\ Rank” summarizes each model’s mean rank across all seven metrics.}
\label{tab:lpips}
\scalebox{0.82}{
\begin{tabular}{lcccccccc}
\toprule
\textbf{Model} 
  & Reflectance $\downarrow$ 
  & Spectral $\downarrow$ 
  & Spatial $\downarrow$ 
  & Synthesis $\uparrow$ 
  & Hallucination $\downarrow$ 
  & Omission $\downarrow$ 
  & Improvement $\uparrow$ 
  & Avg.\ Rank \\
\midrule
ldm\_baseline        
  & 0.1239 ± 0.0405 
  & 12.8441 ± 2.7508 
  & 0.0717 ± 0.0683 
  & 0.0409 ± 0.0290 
  & 0.4558 ± 0.2932 
  & 0.3558 ± 0.2518 
  & 0.1884 ± 0.1232 
  & 6.00 \\
opensrmodel          
  & 0.0076 ± 0.0046 
  &  1.9739 ± 1.0507 
  & 0.0118 ± 0.0108 
  & 0.0194 ± 0.0120 
  & \second{0.0642 ± 0.0271} 
  & 0.6690 ± 0.1291 
  & 0.2668 ± 0.1071 
  & 5.00 \\
satlas               
  & 0.1197 ± 0.0233 
  & 15.1521 ± 2.9876 
  & 0.2766 ± 0.0741 
  & \best{0.0648 ± 0.0302} 
  & 0.5999 ± 0.2182 
  & \best{0.0588 ± 0.0552} 
  & 0.3413 ± 0.1858 
  & 5.00 \\
sr4rs                
  & 0.0979 ± 0.0509 
  & 22.4905 ± 2.1168 
  & 1.0099 ± 0.0439 
  & \second{0.0509 ± 0.0237} 
  & 0.3417 ± 0.1833 
  & \second{0.1924 ± 0.1402} 
  & 0.4659 ± 0.1448 
  & 5.00 \\
superimage           
  & \best{0.0068 ± 0.0016} 
  &  1.8977 ± 1.1053 
  & \best{0.0004 ± 0.0032} 
  & 0.0130 ± 0.0073 
  & \best{0.0357 ± 0.0200} 
  & 0.8844 ± 0.0391 
  & 0.0800 ± 0.0301 
  & 4.28 \\
DiffFuSR (NAIP Harm) 
  & \second{0.0069 ± 0.0031} 
  & \best{1.6738 ± 0.8602} 
  & 0.0060 ± 0.0097 
  & 0.0237 ± 0.0120 
  & 0.1149 ± 0.1020 
  & 0.3466 ± 0.1493 
  & \second{0.5383 ± 0.1240} 
  & \best{3.00} \\
DiffFuSR (NAIP Unharm) 
  & 0.0109 ± 0.0030 
  &  2.0734 ± 1.0205 
  & 0.0061 ± 0.0096 
  & 0.0245 ± 0.0118 
  & 0.1434 ± 0.1158 
  & 0.2846 ± 0.1508 
  & \best{0.5720 ± 0.1242} 
  & \second{3.85}\\
DiffFuSR (Worldstrat)
  & 0.0075 ± 0.0031 
  & \second{1.7327 ± 0.8545} 
  & \second{0.0042 ± 0.0086} 
  & 0.0205 ± 0.0105 
  & 0.0752 ± 0.0682 
  & 0.6216 ± 0.1214 
  & 0.3032 ± 0.0878 
  & \second{3.85} \\

\bottomrule
\end{tabular}
}
\end{table*}

\subsubsection{Fusion-based enhancement of multispectral bands}

To super-resolve all 12 Sentinel-2 bands, we combine RGB upsampling with learned fusion model\del{s} trained to upscale 10 m, 20 m, and 60 m bands to a unified 2.5 m resolution. Table~\ref{tab:fineperf} presents a comparison with classical \add{baselines and the FusionNet baseline \mbox{\cite{deng2022machine}}}\del{GS pansharpening}. \del{Across all data sources, our neural network (NN)-based fusion dramatically outperforms GS in terms of ERGAS, PSNR, and SSIM. For instance, ERGAS drops from over $2900$ (GS) to under $95$ (NN), PSNR increases from ~22 dB to ~32 dB, and SSIM rises from ~0.65 to ~0.88.}\add{Since HR ground truth is not available for all bands, performance must be assessed indirectly. To this end, we downsample the super-resolved outputs back to the native Sentinel-2 resolution and compare them there. Overall, our approach outperforms all classical baselines except for the GLP method~\cite{vivone2018full}, which remains highly competitive, particularly on the 20\,m and 60\,m bands (see Table~\ref{tab:fineperf}).}

\showonly{
\begin{table*}[htbp]
\centering
\caption{\del{Quantitative Comparison of Super-Resolution Fusion Methods (NN vs. GS) for 12 Sentinel-2 Bands (Best Across Dataset is \textbf{Bold})}}
\begin{tabular}{|l|l|c|c|c|c|c|c|}
\hline
\textbf{Data Source} & \textbf{Method} & \textbf{R\textsuperscript{2}} & \textbf{Cross-Correlation} & \textbf{SSIM} & \textbf{PSNR} & \textbf{MSE} & \textbf{ERGAS} \\
\hline
\multirow{2}{*}{NAIP Harmonized } 

& GS & -0.343 & 0.905 & 0.648 & 22.76 & 1002.61 & 2972.76 \\
& NN & 0.804 &\textbf{0.981} & 0.877 & \textbf{32.34} & 143.01 & 94.83 \\
\hline
\multirow{2}{*}{NAIP Unharmonized } 

& GS & -0.306 & 0.908 & 0.656 & 22.95 & 977.47 & 3074.05 \\
& NN & 0.805 & \textbf{0.981} & 0.876 & 32.10 & 142.45 & 95.42 \\
\hline

\multirow{2}{*}{WorldStrat } 

& GS & -0.341 & 0.905 & 0.648 & 22.75 & 1001.48 & 57515.33 \\
& NN & \textbf{0.806} & \textbf{0.981} & \textbf{0.878} & 32.27 & \textbf{141.96} & \textbf{94.68} \\
\hline
\end{tabular}
\label{tab:sr_fusion_comparison}
\end{table*}
}

\begin{table}[h!]
\centering
\small
\setlength{\tabcolsep}{6pt}
\renewcommand{\arraystretch}{1.2}
\caption{\add{Performance of the WorldStrat-trained super-resolution model coupled with our proposed GLP-NN fusion model, evaluated on bands at 10 m, 20 m, and 60 m resolution. The baselines include Generalized Laplacian Pyramid (GLP) \cite{vivone2018full}, Gram Schmidt (GS) \cite{dalla2015global}, FusionNet \cite{deng2022machine}, \best{BOLD} indicates best and \second{UNDERLINED} indicates second best}}
\label{tab:fineperf}
\scalebox{0.73}{ 
\begin{tabular}{llcccccc}
\hline
\textbf{Resolution} & \textbf{Method} & \textbf{PSNR} & \textbf{SSIM} & \textbf{MSE} & \textbf{R$^2$} & \textbf{NCC} & \textbf{ERGAS} \\
\hline
\multirow{6}{*}{\makecell{10m \\ (Ratio 4)} } 
 & GLP             & \second{27.95} & \second{0.85}2 & \second{12.08}  & \second{0.977} & \second{0.997} & \second{2.766} \\
 & \textit{GLP NN}       & \best{29.57} & \best{0.904} & \best{8.56}   & \best{0.984} & \best{0.998} & \best{2.260} \\
 & IHS            & 27.90 & 0.870 & 15.82  & 0.972 & 0.996 & 2.583 \\
 & GS             & 16.75 & 0.678 & 188.99 & 0.717 & 0.976 & 10.141 \\
 & PCA            & 17.53 & 0.118 & 254.67 & 0.552 & 0.941 & 10.627 \\
 & Fusion Net     & 29.48 & 0.898 & 8.62   & 0.984 & \second{0.998} & 2.275 \\
\hline
\multirow{6}{*}{\makecell{20m \\ (Ratio 8)} } 
 & GLP            & \best{24.39} & \best{0.836} & \best{15.51}  & \best{0.952} & \best{0.998} & \best{0.789} \\
 & \textit{GLP NN}         & \second{22.01} & \second{0.757} & \second{26.26}  & \second{0.916} & \second{0.997} & \second{1.030} \\
 & IHS            & 17.92 & 0.571 & 84.89  & 0.782 & 0.990 & 1.820 \\
 & GS             & 4.38  & 0.442 & 1584.73& -3.885& 0.908 & 7.772 \\
 & PCA            & 15.09 & 0.195 & 257.04 & 0.295 & 0.970 & 2.948 \\
 & Fusion Net     & 21.10 & 0.767 & 32.86  & 0.904 & \second{0.997} & 1.288 \\
\hline
\multirow{6}{*}{\makecell{60m \\ (Ratio 24)} } 
 & GLP            & \best{27.92} & \best{0.957} & \best{3.19}   & \best{0.997} & \best{0.999} & \best{0.188} \\
 & \textit{GLP NN}         & \second{23.65} & \second{0.909} & \second{5.05}   & \second{0.995} & \second{0.999} & \second{0.322} \\
 & IHS            & 22.63 & 0.828 & 21.25  & 0.982 & 0.997 & 0.312 \\
 & GS             & 7.42  & 0.515 & 847.22 & 0.310 & 0.937 & 1.937 \\
 & PCA            & 17.79 & 0.473 & 86.33  & 0.926 & 0.987 & 0.689 \\
 & Fusion Net     & 23.16 & 0.918 & 7.21   & 0.992 & \second{0.999} & 0.380 \\
\hline
\end{tabular}
}
\end{table}

\add{\subsection{Ablations}}
\add{Our experiments show that harmonization provides a consistent, albeit modest, improvement in performance.}

\begin{figure}[htbp]
  \centering

  \begin{minipage}[t]{0.151\textwidth}
    \centering
    \includegraphics[width=\linewidth, trim=100 100 100 100, clip]{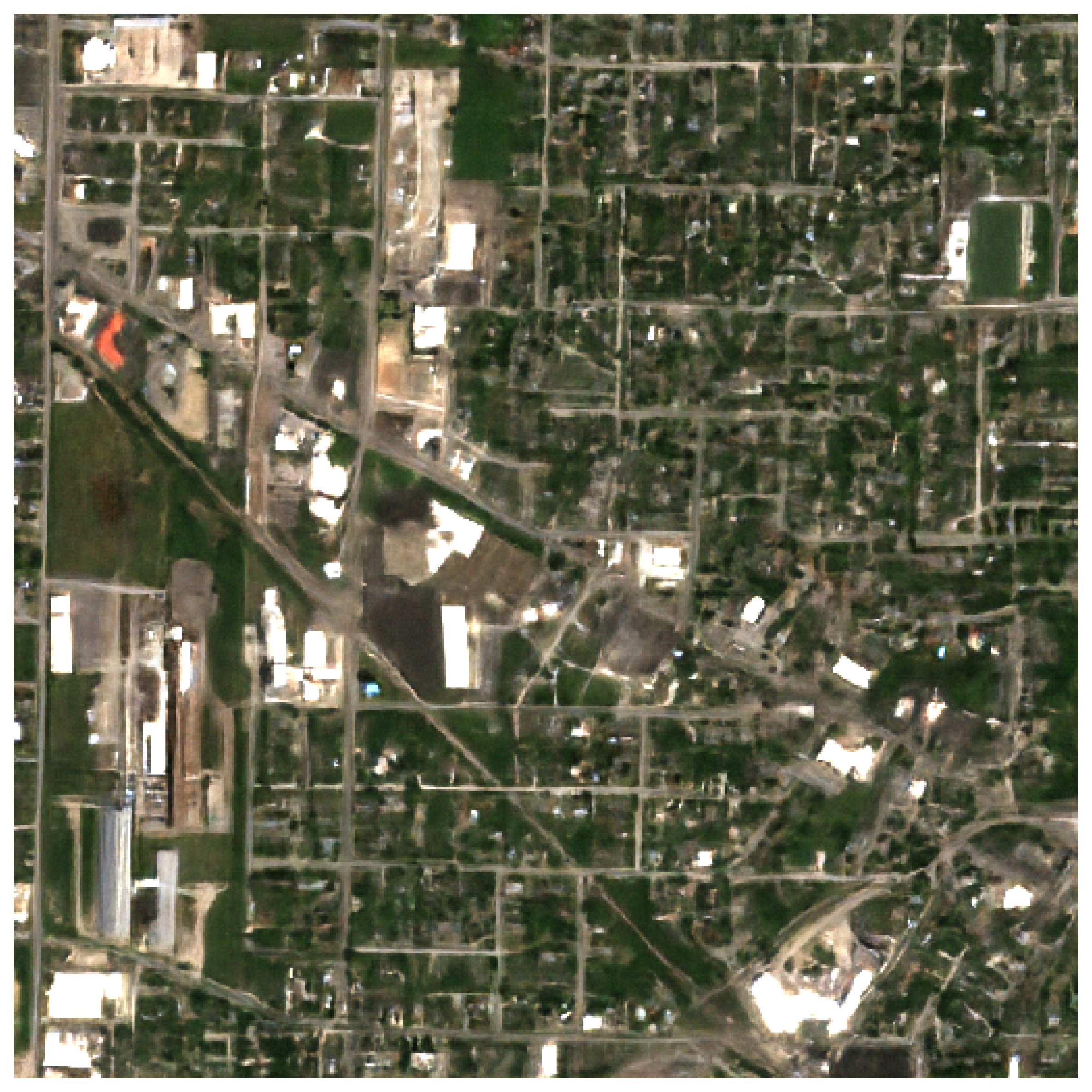}\\
    \small NAIP Harmonized
  \end{minipage}\hfill
  \begin{minipage}[t]{0.151\textwidth}
    \centering
    \includegraphics[width=\linewidth ,trim=100 100 100 100, clip]{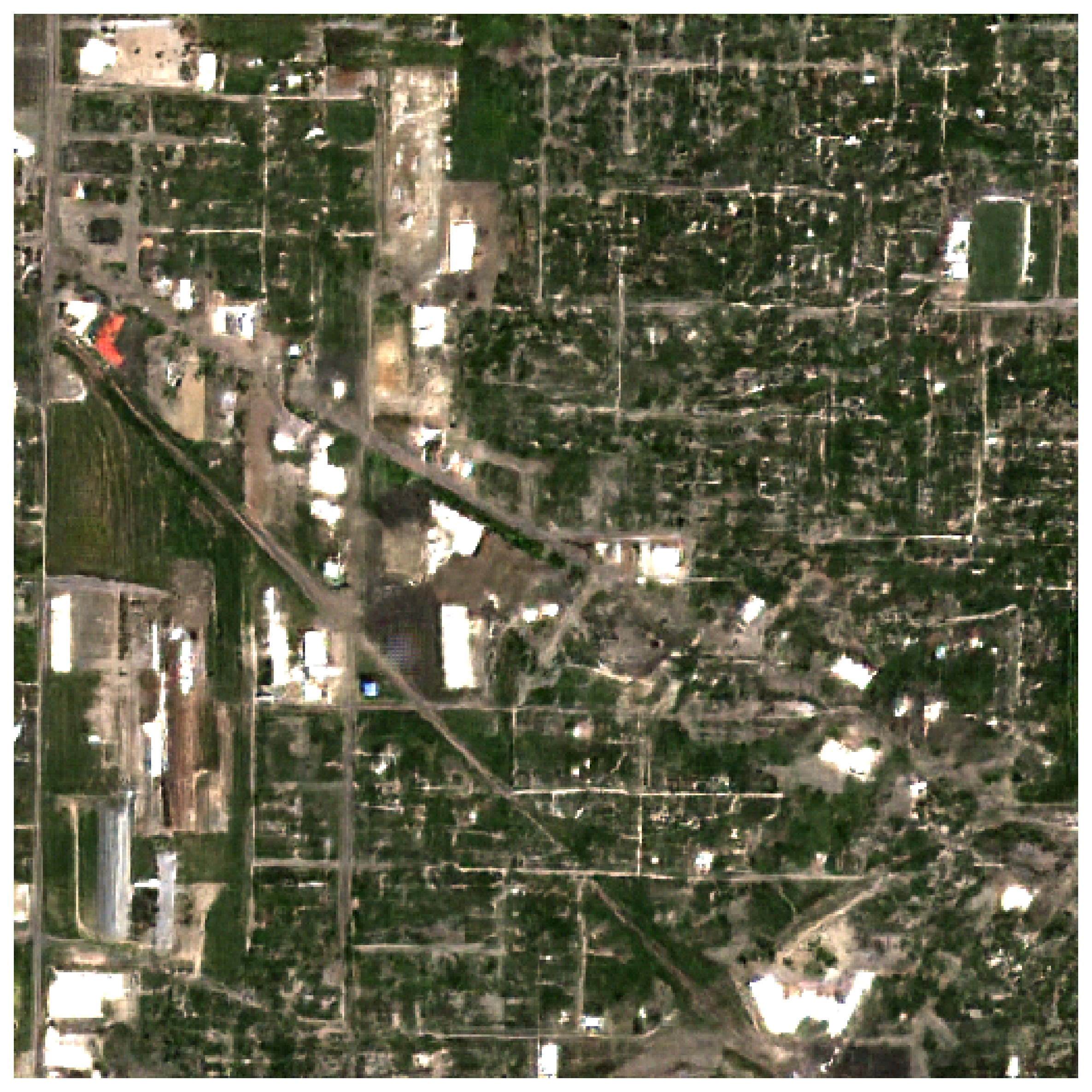}\\
    \small NAIP UnHarmonized
  \end{minipage}\hfill
  \begin{minipage}[t]{0.151\textwidth}
    \centering
    \includegraphics[width=\linewidth, trim=100 100 100 100, clip]{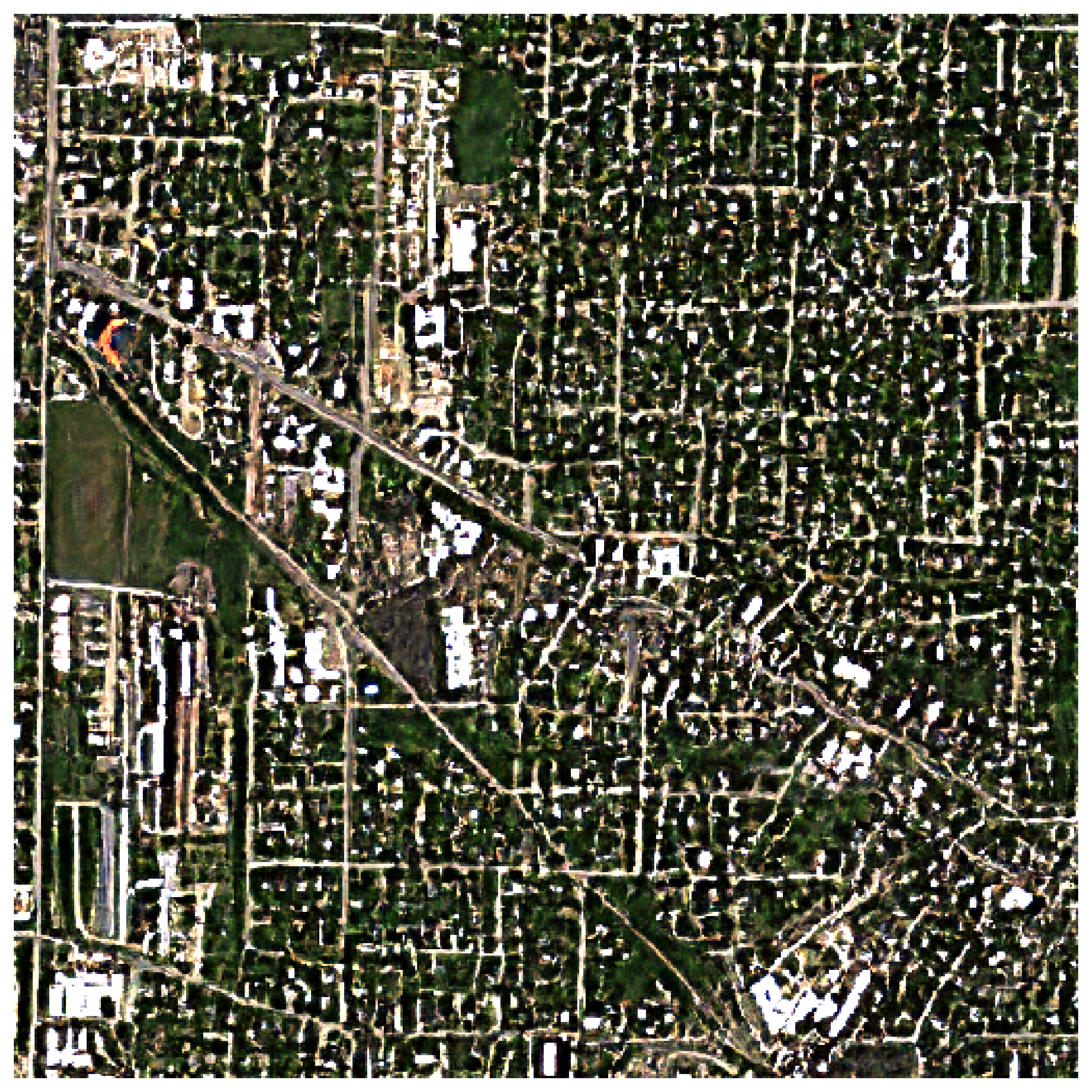}\\
    \small NAIP Harm, Fixed Kernel
  \end{minipage}

  \caption{\add{Visual comparison of Ablations of our SR method on OpenSR-test ~\cite{opensrtest} test image. Among all methods, DiffFuSR with NAIP harmonized produces the sharpest reconstruction with fewer artifacts, using fixed kernel instead of blind kernel severly degrades the results.}}
  \label{fig:sr_ablation_roi0117}
\end{figure}

\add{A more critical factor is the choice of the degradation kernel. Instead of fixing the Gaussian blur kernel to a single value (e.g., $\sigma=3.0$), we replace it with a \emph{blind kernel} sampled across a range ($\sigma \in [2,4]$). This change produces a dramatic improvement: with a fixed kernel the model performance degrades sharply, while the blind kernel yields much more robust results. The importance of this choice is evident both quantitatively (Table~\ref{tab:naip_ablation_all}) and qualitatively (see Figure~\ref{fig:sr_ablation_roi0117}). Without the blind kernel, the entire pipeline nearly collapses, highlighting its central role in the training setup.
}

\begin{table*}[t]
\centering
\caption{\add{Ablation metrics across LPIPS, Normalized Difference, and CLIP. (↓ lower is better; ↑ higher is better). The selected models for comparison are trained on NAIP data using three setting i.e. Harmonized data with anisotropic blind kernel(Harm), Unharmonized data without harmonization and Harmonized data with a fixed kernel}}
\label{tab:naip_ablation_all}
\scalebox{0.72}{
\begin{tabular}{lccc ccc ccc}
\toprule
& \multicolumn{3}{c}{\textbf{LPIPS}} & \multicolumn{3}{c}{\textbf{Normalized Difference}} & \multicolumn{3}{c}{\textbf{CLIP}} \\
\cmidrule(lr){2-4}\cmidrule(lr){5-7}\cmidrule(lr){8-10}
\textbf{Metric} & \textbf{Harm} & \textbf{Unharm} & \textbf{Harm (Fix Kernel)} & \textbf{Harm} & \textbf{Unharm} & \textbf{Harm (Fix Kernel)} & \textbf{Harm} & \textbf{Unharm} & \textbf{Harm (Fix Kernel)} \\
\midrule
Reflectance $\downarrow$
& $0.0069 \pm 0.0031$ & $0.0109 \pm 0.0030$ & $0.0329 \pm 0.0187$
& $0.0024 \pm 0.0009$ & $0.0038 \pm 0.0008$ & $0.0122 \pm 0.0067$
& $0.0069 \pm 0.0032$ & $0.0109 \pm 0.0030$ & $0.0329 \pm 0.0187$ \\
Spectral $\downarrow$
& $1.6738 \pm 0.8602$ & $2.0734 \pm 1.0205$ & $7.3211 \pm 3.1621$
& $1.1103 \pm 0.6197$ & $1.2320 \pm 0.7256$ & $1.5652 \pm 0.9009$
& $1.6739 \pm 0.8602$ & $2.0734 \pm 1.0205$ & $7.3211 \pm 3.1621$ \\
Spatial $\downarrow$
& $0.0060 \pm 0.0097$ & $0.0061 \pm 0.0096$ & $0.0132 \pm 0.0123$
& $0.0052 \pm 0.0092$ & $0.0053 \pm 0.0093$ & $0.0130 \pm 0.0118$
& $0.0061 \pm 0.0098$ & $0.0061 \pm 0.0096$ & $0.0132 \pm 0.0123$ \\
Synthesis $\uparrow$
& $0.0237 \pm 0.0120$ & $0.0245 \pm 0.0118$ & $0.0570 \pm 0.0325$
& $0.0083 \pm 0.0043$ & $0.0086 \pm 0.0042$ & $0.0187 \pm 0.0101$
& $0.0238 \pm 0.0120$ & $0.0245 \pm 0.0118$ & $0.0570 \pm 0.0325$ \\
Hallucination $\downarrow$
& $0.1149 \pm 0.1020$ & $0.1434 \pm 0.1158$ & $0.5222 \pm 0.1747$
& $0.4580 \pm 0.1020$ & $0.4927 \pm 0.1095$ & $0.7463 \pm 0.0521$
& $0.2883 \pm 0.2333$ & $0.3272 \pm 0.2400$ & $0.3810 \pm 0.1223$ \\
Omission $\downarrow$
& $0.3466 \pm 0.1493$ & $0.2846 \pm 0.1508$ & $0.0508 \pm 0.0368$
& $0.3543 \pm 0.0836$ & $0.3373 \pm 0.0870$ & $0.1376 \pm 0.0373$
& $0.3358 \pm 0.2214$ & $0.2929 \pm 0.2169$ & $0.1003 \pm 0.0289$ \\
Improvement $\uparrow$
& $0.5383 \pm 0.1240$ & $0.5720 \pm 0.1242$ & $0.4269 \pm 0.1611$
& $0.1876 \pm 0.0344$ & $0.1701 \pm 0.0351$ & $0.1160 \pm 0.0207$
& $0.3759 \pm 0.1444$ & $0.3799 \pm 0.1475$ & $0.5186 \pm 0.1202$ \\
\bottomrule
\end{tabular}
}
\end{table*}

\section{Discussion}

\begin{figure}[htbp]
  \centering

  \begin{minipage}[t]{0.151\textwidth}
    \centering
    \includegraphics[width=\linewidth, trim=100 100 100 100, clip]{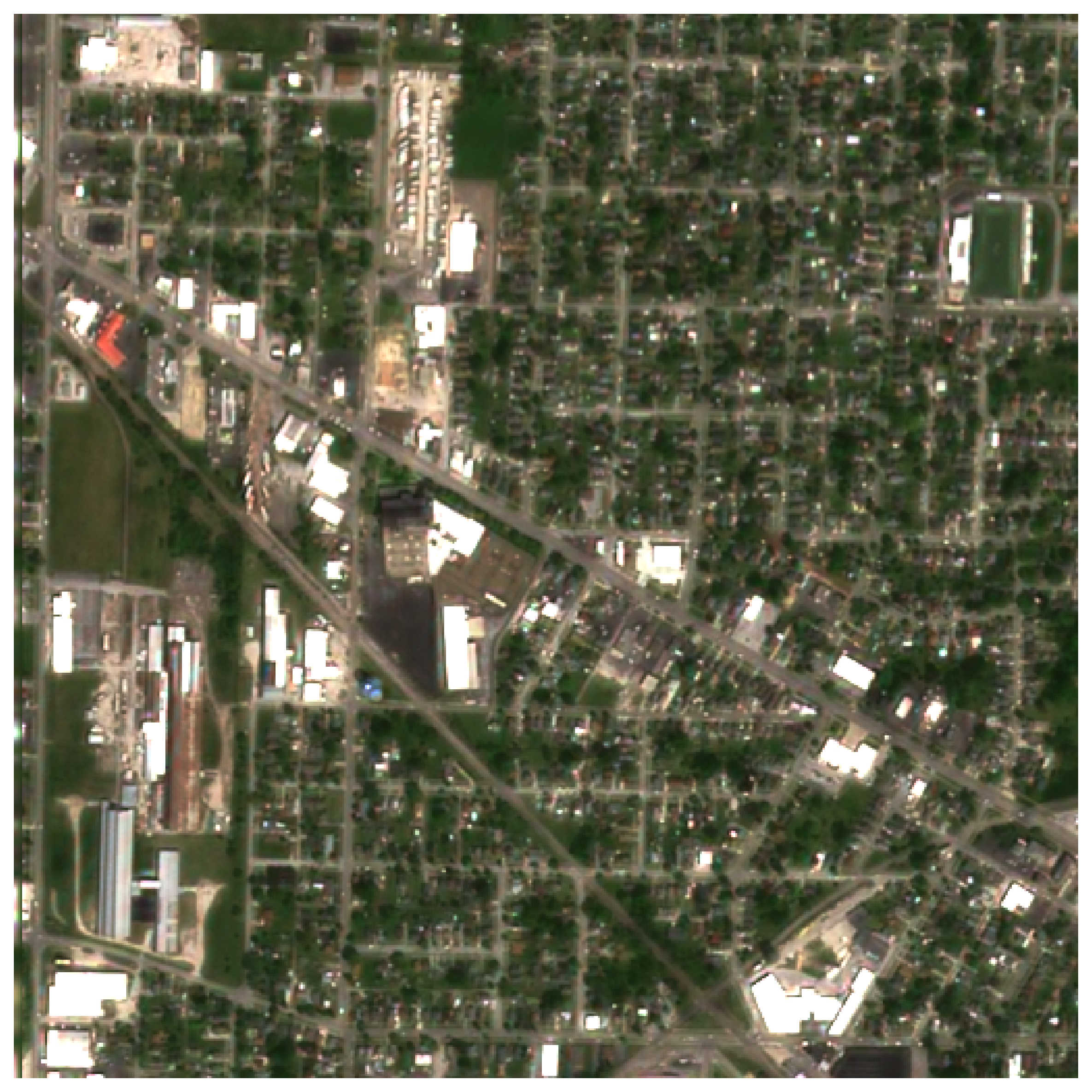}\\
    \small HR (Reference)
  \end{minipage}\hfill
  \begin{minipage}[t]{0.151\textwidth}
    \centering
    \includegraphics[width=\linewidth, trim=100 100 100 100, clip]{figures/comparison/ROI_0117__20170515T162341_20170515T163103_T17TLF_naip_harm_true_color.png}\\
    \small DiffFuSR (NAIP Harmonized)
  \end{minipage}\hfill
  \begin{minipage}[t]{0.151\textwidth}
    \centering
    \includegraphics[width=\linewidth, trim=100 100 100 100, clip]{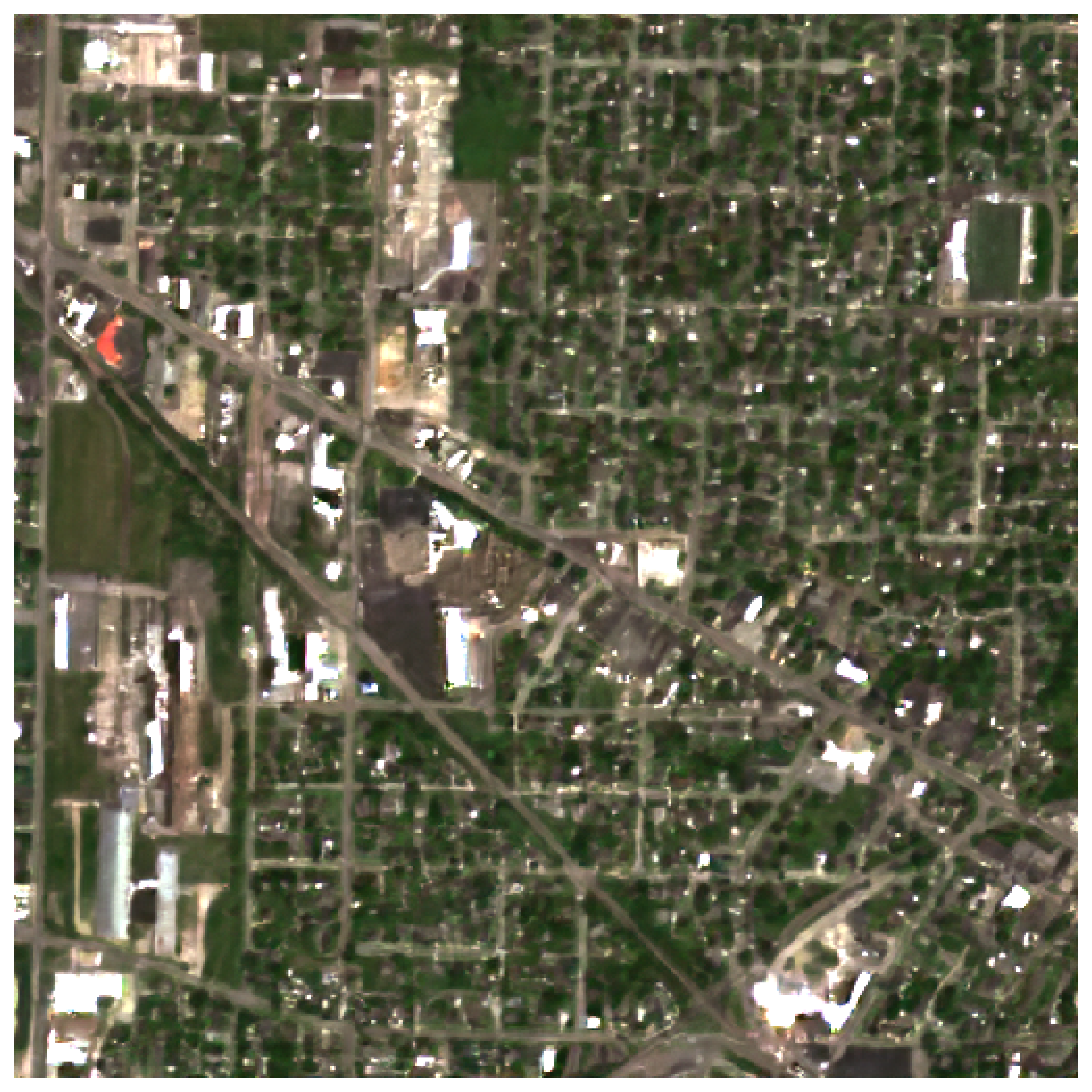}\\
    \small OpenSR
  \end{minipage}

  \vspace{1em}

  \begin{minipage}[t]{0.151\textwidth}
    \centering
    \includegraphics[width=\linewidth, trim=100 100 100 100, clip]{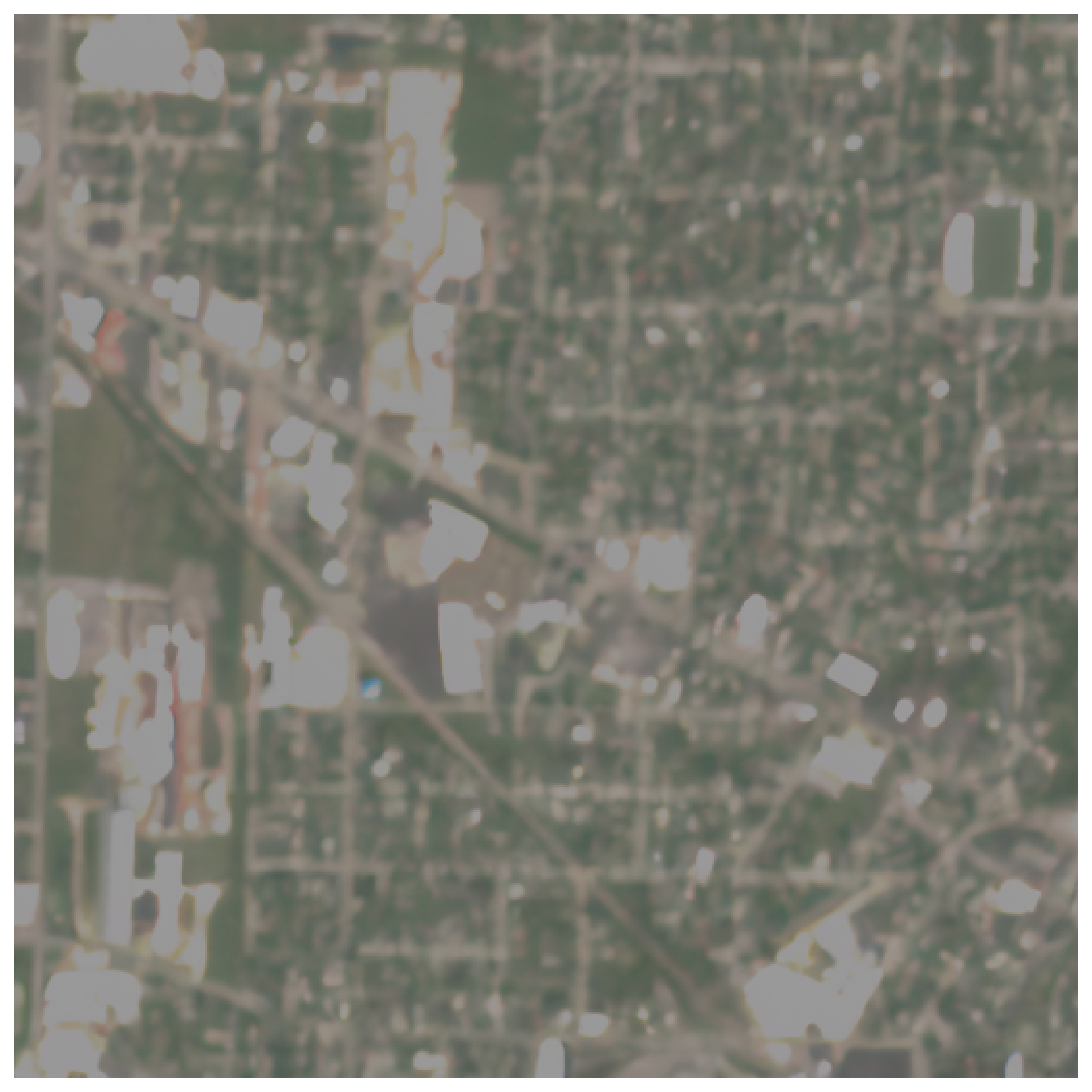}\\
    \small LDM\_Baseline
  \end{minipage}\hfill
  \begin{minipage}[t]{0.151\textwidth}
    \centering
    \includegraphics[width=\linewidth, trim=100 100 100 100, clip]{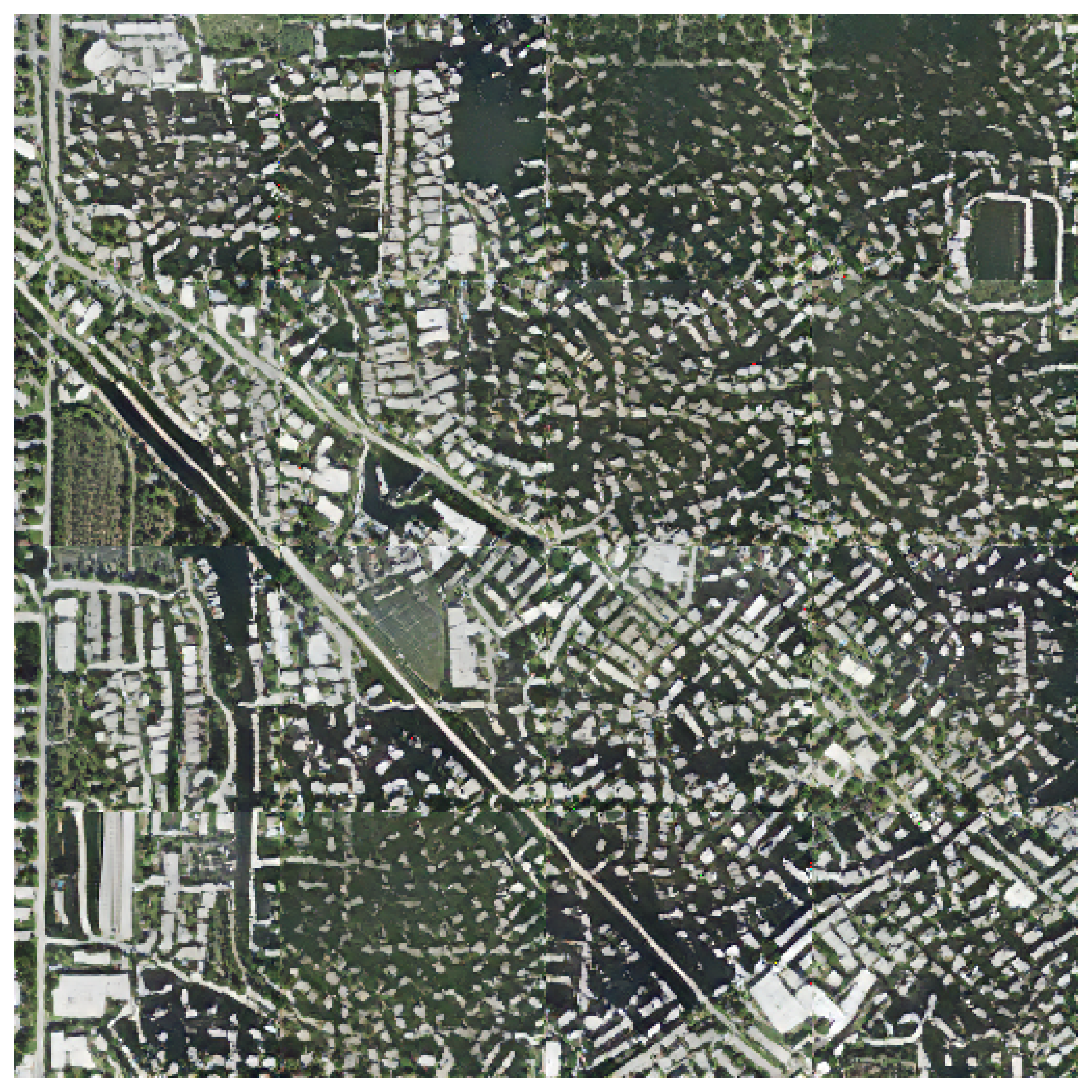}\\
    \small Satlas
  \end{minipage}\hfill
  \begin{minipage}[t]{0.151\textwidth}
    \centering
    \includegraphics[width=\linewidth, trim=100 100 100 100, clip]{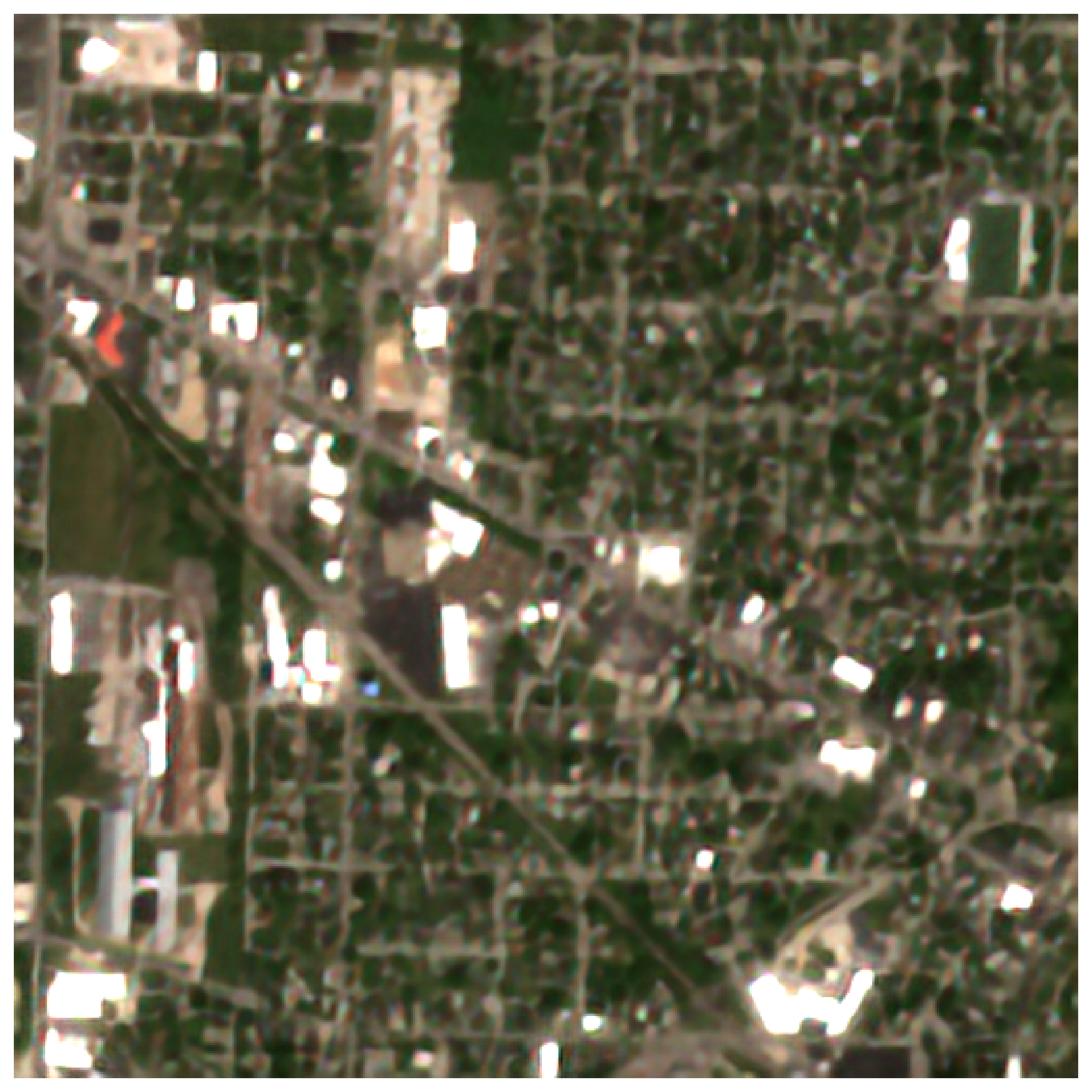}\\
    \small SuperImage
  \end{minipage}

  \caption{
    Visual comparison of SR methods on an OpenSR-test~\cite{opensrtest} image.
    The top row highlights the strongest models: HR reference, our model trained on harmonized NAIP, and OpenSR.
    The bottom row shows the remaining baselines.
    Among all methods, DiffFuSR produces the sharpest reconstruction with fewer artifacts,
    while OpenSR remains a strong baseline.
    LDM\_Baseline and Satlas show degraded outputs, and SuperImage tends to oversmooth fine structures.
  }
  \label{fig:sr_comparison_roi0117}
\end{figure}

\del{We presented DiffFuSR, a modular pipeline to super-resolve all 12 spectral bands of Sentinel-2 imagery to a 2.5 m GSD, leveraging a diffusion-based model for RGB SR and a learned network for multispectral fusion. The evaluation of the proposed RGB SR model against several contemporary baselines on the OpenSR-test benchmark demonstrated its capabilities in quantitative terms. However, the qualitative results need to be paired with quantitative results to accurately gauge the true performance. } \add{We note that  the qualitative results need to be paired with quantitative results to accurately gauge the true performance.} Looking at one of the columns of results in quantitative Tables can be misleading, e.g., Satlas and LDM\_Baseline seem to be doing the best in terms of Omission metric in Table~\ref {tab:lpips}. But when looking at the rest of the values and also the qualitative results in Fig.~\ref {fig:sr_comparison_roi0117}, the SR outputs contain severe artefacts. 

\del{When trained with harmonized NAIP data, the proposed diffusion model consistently demonstrated superior or highly competitive performance across multiple quantitative metrics, including Normalized Difference, CLIP distance, and LPIPS (Tables \ref{tab:nd} - \ref{tab:lpips}). For instance, it achieved the lowest reflectance error (0.0024) and highest spectral fidelity (1.1103) in Normalized Difference metrics, and the best spectral score (1.6739) with CLIP distance. It also recorded a competitive hallucination score (0.1149) using LPIPS, alongside one of the best improvement scores (0.5383) with the same metric, indicating an enhanced ability to reconstruct rich, meaningful high-frequency content with minimal introduction of spurious details.} 
\add{The strong performance of the harmonized NAIP model across all metric categories (Tables \ref{tab:nd} - \ref{tab:lpips}) highlights the critical role of spectral harmonization. By aligning the source domain (NAIP) with the target (Sentinel-2), the model achieves superior reflectance and spectral fidelity, indicating it learns a more accurate representation of Sentinel-2's characteristics.}
Qualitatively, particularly the harmonized NAIP variant, produced sharper reconstructions with fewer artifacts compared to strong baselines like OpenSR, and markedly better results than LDM\_Baseline, Satlas, or Superimage, which exhibited issues such as instability, poor spectral integrity, or over-smoothing, respectively (Fig.~\ref{fig:sr_comparison_roi0117}). 

Traditional non-generative SR approaches often optimize for pixel-wise fidelity, which can result in perceptually subdued images. While some modern regression-based models like Superimage offer good reflectance fidelity, our generative approach aims for, and achieves, a compelling balance of perceptual realism and quantitative accuracy. In the domain of generative models, GAN-based approaches like Satlas showed a tendency for poorer spectral integrity despite visual sharpness. Our diffusion model, benefiting from robust training strategies, including the blind degradation modeling, provides a more controlled generation process.


\del{For the subsequent enhancement of all 12 Sentinel-2 bands, the learned multispectral fusion network was evaluated against \del{classical GS pansharpening}\add{multiple baselines including}. The results \add{show that} \del{were definitive:} the \add{proposed} neural network (\add{GLP-}NN)-based fusion outperformed these baselines \del{GS}\del{baselines} across \add{most}\del{all} considered metrics (R\^2, Cross-Correlation, SSIM, PSNR, MSE, and ERGAS) \del{and for all RGB SR inputs} (Tab.~\ref{tab:fineperf})\add{, except for the GLP baseline, which performs significantly better}} \add{However, the superior performance of the GLP baseline when matching low-resolution images does not translate into qualitative results. As shown in Fig.~\ref{fig:fusion}, the GLP baseline produces very blurry outputs, whereas our proposed method generates sharp results with well-preserved details while remaining the most accurate in terms of quantitative metrics}. \del{For example, when using the harmonized NAIP SR RGB, ERGAS values dropped from over 2900 for GS to under 95 for our NN approach, and PSNR increased from approximately 22 dB to over 32 dB. Qualitative comparisons (Fig.~\ref{fig:fusion}) further confirmed that our fusion model better preserves semantic structures and avoids the spectral distortions common in GS fusion, especially for the 20 m and 60 m bands.}

A critical factor in the success of the RGB SR stage was the spectral harmonization and degradation modeling applied to the proxy HR datasets (NAIP and WorldStrat) to simulate Sentinel-2 characteristics. \del{The superior performance of models trained on harmonized NAIP data, particularly for vegetated scenes (Fig. \ref{fig:rgbSR1}), underscores the importance of these pre-processing steps.} The blind degradation modeling, facilitated by a contrastive degradation encoder, is a significant contribution, enabling the SR model to adapt to unseen degradation conditions in real Sentinel-2 imagery without explicit kernel information.

\del{The choice of HR training data also influenced performance, with NAIP-trained models excelling in natural, vegetated environments, while WorldStrat-trained models demonstrated an advantage in structured, urban settings (Figs.~\ref{fig:rgbSR1} and \ref{fig:rgbSR2}).} This suggests that while harmonization is effective, the inherent spatial and radiometric properties of the proxy data can still subtly influence model specialization, highlighting the value of diverse training datasets for developing globally applicable models.

\del{The effect of harmonization on NAIP data is most prominent on reflectance and spectral metrics. Other metrics are not affected to the same extent. This is in accordance to our expectation as harmonization brings the spectral content of NAIP closer to Sentinel-2. It is also interesting to note that metrics like hallucination are also slightly reduced, giving interesting insight into the effect of harmonization. On the other hand, the omission metric is actually better in the unharmonized case. This behavior is interesting because it indicates that the SR images have missing details that were present in the HR image.} 

The learned fusion network, trained using only native Sentinel-2 data via a Wald protocol simulation, proved highly effective. Its ability to utilize the 2.5 m super-resolved RGB output from the diffusion model and to upscale the 10 m, 20 m, and 60 m bands while preserving spectral integrity is a core achievement. The scale-aware design, with dedicated fusion modules and tailored inputs for each Sentinel-2 resolution group (10 m, 20 m, 60 m bands), contributes to the robustness and high fidelity of the final super-resolved Sentinel-2 12-band product. This modularity, both within the fusion stage and in the overall two-stage pipeline, offers practical benefits for future refinement and adaptation. There is room for improvement in the design choices in the Fusion module architecture\add{, to match the quantitative performance of the GLP baseline}, which we leave for future work.
\add{\subsection{Implications of the Evaluation Strategy}}
\add{A key strength of this study's SR evaluation is the use of the OpenSR-test benchmark. By assessing performance across three different distance metrics (Normalized Difference, LPIPS, and CLIP), we can move beyond simple pixel error. This provides a more holistic validation, capturing physical accuracy (reflectance), perceptual realism (LPIPS), and semantic consistency (CLIP), which is critical for gauging the true utility of generative models.}

\add{It is crucial to properly interpret the quantitative results for the multispectral fusion (Tab.~\ref{tab:fineperf}). In the absence of high-resolution ground truth for all 12 Sentinel-2 bands, performance was assessed indirectly by downsampling the super-resolved outputs back to their native resolution and comparing them there. This "no-reference" validation, based on the Wald protocol, primarily measures fidelity to the original low-resolution signal, not absolute accuracy at the 2.5 m target resolution. This limitation explains the apparent discrepancy noted in our results: the classical GLP method excelled in the quantitative metrics (by producing a blurry result very similar to the downsampled input) but was clearly inferior in qualitative assessment (Fig.~\ref{fig:fusion}), where it produces very blurry outputs. Our proposed GLP-NN, therefore, strikes a better balance, enhancing spatial detail while remaining acceptably consistent with the original spectral information.}

\add{
\section{Limitations and Future Work}}

\add{While DiffFuSR advances Sentinel-2 super-resolution, several limitations remain that can be addressed in future work:}

\begin{itemize}
    \item \add{\textbf{Proxy supervision and domain gap:} Training the RGB SR model on NAIP and WorldStrat data with statistics-based harmonization cannot fully remove sensor and illumination differences. Residual domain gaps may bias colors and textures in super-resolved outputs. \textit{Future research could explore large-scale datasets of harmonized cross-sensor imagery, such as extended SEN2NAIP \cite{aybar2024sen2naip} variants with more temporal diversity, which could help model real-world variations more effectively.}}

    \item \add{\textbf{Guidance bias in fusion:} Using SR-RGB as a spatial prior can transfer RGB-specific artifacts or impose structures not present in non-RGB bands (e.g., SWIR), risking subtle spectral distortions despite our safeguards. \textit{Future work may investigate fusion models with adaptive spectral weighting to dynamically regulate RGB influence per band.}}

    \item \add{\textbf{Evaluation constraints:} In the absence of HR Sentinel-2 ground truth, we rely on OpenSR-style proxies and Wald-based no-reference validation; these may not perfectly reflect absolute radiometric fidelity in the wild. \textit{Future evaluation can come up with suitable downstream applications that demonstrate the utility of the super-resolved bands}}
\end{itemize}

\section{Conclusions}

In this work, we proposed and demonstrated DiffFuSR, a pipeline to super-resolve all 12 bands of Sentinel-2. The main contribution of the work was the combination of strengths of both classical and state-of-the-art deep learning methods, and ensuring best practices for pre-processing and harmonizing the data. 
The work has demonstrated that learning state-of-the-art Sentinel-2 SR models from other data modalities, e.g., NAIP, is not only possible but one of the best ways due to the lack of paired data. Converting the NAIP data to look like Sentinel-2 is the most critical factor in the pipeline. Small details like the correct kernel size and the type of blur kernel are crucial factors for an accurate degradation model to simulate Sentinel-2 images. \del{By learning from NAIP data with the correct degradation model, we were able to train a diffusion model to super-resolve the RGB bands of Sentinel-2. To super-resolve all 12 bands, the super-resolved RGB image was used to guide the enhancement of the remaining bands in the proposed fusion module.} 
\add{\del{The proposed fusion module is a neural network inspired by the classical pansharpening paradigm, particularly the MTF-GLP framework.} By embedding the principles of modulation transfer function filtering and regression-based detail injection into a learnable architecture, the model inherits the interpretability and stability of traditional approaches while benefiting from the adaptability of deep learning. \del{This hybrid design enables consistent spectral preservation and enhanced spatial detail, outperforming existing neural fusion methods in both quantitative metrics and visual quality.}}

\del{However, some limitations and areas for future development should be acknowledged. The computational requirements for training and, more critically, for inference with diffusion models are substantial due to their iterative sampling nature. For widespread operational adoption, research into model compression, knowledge distillation, or more efficient sampling strategies specifically for SR tasks will be crucial. While the training datasets (SEN2NAIP, WorldStrat) are extensive, developing models that exhibit even greater robustness across the full diversity of global landscapes and atmospheric conditions will likely benefit from even larger and more varied training corpora, potentially incorporating advanced data augmentation or domain generalization techniques. The reliance on proxy data, even with careful harmonization, may also leave residual domain gap effects that warrant continued research into cross-domain adaptation.}

\del{Future work could extend this framework to incorporate temporal information from Sentinel-2 time series, potentially improving SR quality and enabling consistent 2.5 m analysis-ready data stacks. Adapting the degradation models and fusion strategies for other multispectral satellite sensors could broaden the applicability of this promising pipeline architecture.}

\add{Beyond the technical contributions, DiffFuSR represents a practical step toward bridging the gap between open-access data and costly high-resolution commercial imagery. By providing a reliable method to generate a 2.5 m, 12-band Sentinel-2 product, this work can unlock new capabilities for a wide range of applications, such as detailed land cover mapping, agricultural monitoring, and urban analysis, for researchers who rely solely on the public Copernicus data stream. Ultimately, this work demonstrates that a modular pipeline, thoughtfully combining generative models with classical principles and rigorous data handling, is key to enhancing the value of foundational EO datasets like Sentinel-2.}

\section*{Acknowledgments}
Thanks to the European Space Agency, PRODEX program, for funding
under the project 4000141281.

    \bibliographystyle{IEEEtran}
    \bibliography{sample}

\section{Supplementary Materials}

\begin{figure*}[htbp]
    \centering

    \subfigure{\includegraphics[width=0.320\textwidth]{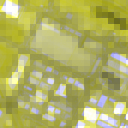}}
    \hfill
    \subfigure{\includegraphics[width=0.320\textwidth]{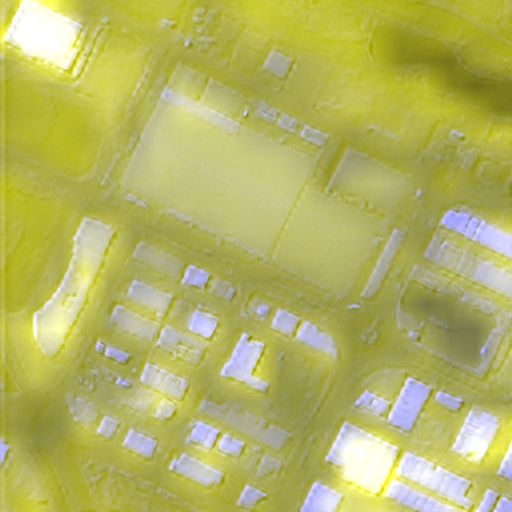}}
    \hfill
    \subfigure{\includegraphics[width=0.320\textwidth]{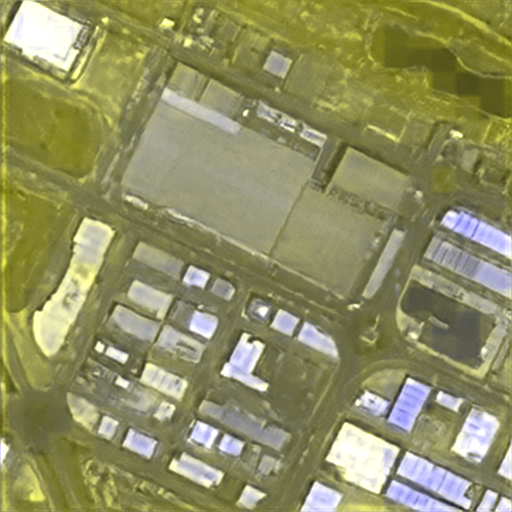}}
    \begin{tabular}{ccc}
        \makebox[0.30\textwidth]{\textbf{Low Resolution}} &
        \makebox[0.30\textwidth]{\textbf{GLP \cite{vivone2018full}}} &
        \makebox[0.30\textwidth]{\textbf{GLP Fusion Network(ours)}}
    \end{tabular}

    \subfigure{\includegraphics[width=0.320\textwidth]{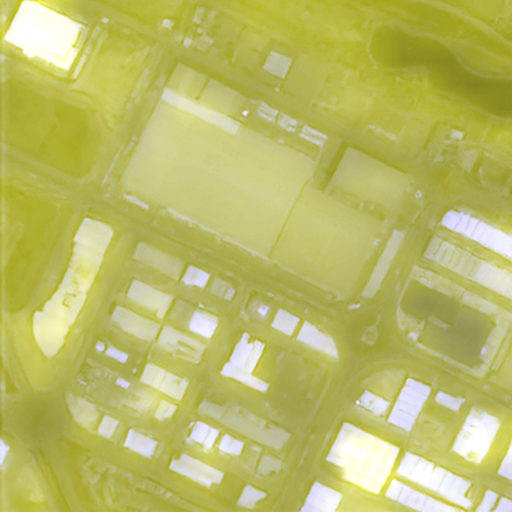}}
    \hfill
    \subfigure{\includegraphics[width=0.320\textwidth]{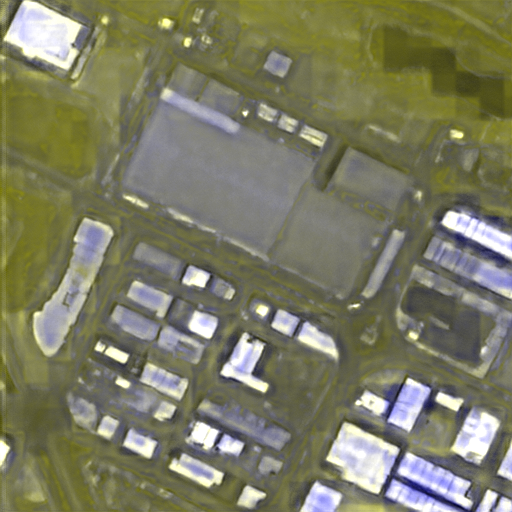}}
    \hfill
    \subfigure{\includegraphics[width=0.320\textwidth]{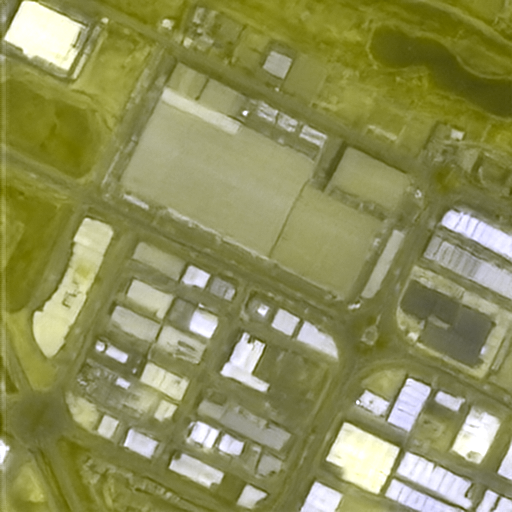}}
    \begin{tabular}{ccc}
        \makebox[0.30\textwidth]{\textbf{IHS}} &
        \makebox[0.30\textwidth]{\textbf{Fusion Network \cite{deng2022machine}}} &
        \makebox[0.30\textwidth]{\textbf{Gram Schmidt \cite{dalla2015global}}}
    \end{tabular}

    \caption{\add{Fusion output comparisons across 60\,m atmospheric bands. The figure show the Low Resolution input, Generalized laplacian pyramid (GLP) baseline method \cite{vivone2018full}, our GLP inspired neural network, intensity-hue-saturation (IHS) transform, and Fusion network baseline \cite{deng2022machine} and  Gram Schmidt \cite{dalla2015global} method.}}

    \label{fig:fusion1}
\end{figure*}

\begin{figure*}[htbp]
    \centering 
    
    \subfigure{\includegraphics[width=0.320\textwidth]{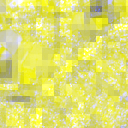}}
    \hfill
    \subfigure{\includegraphics[width=0.320\textwidth]{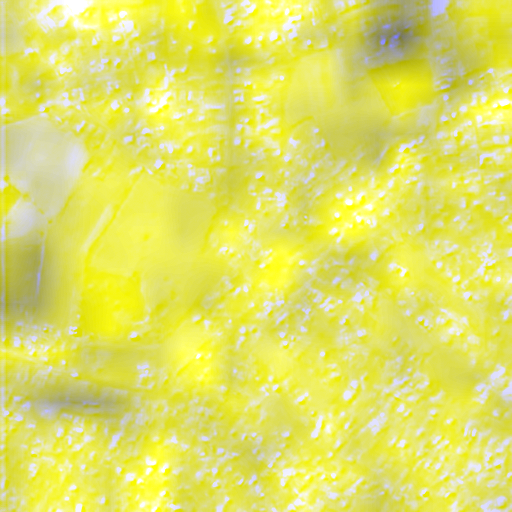}}
    \hfill
    \subfigure{\includegraphics[width=0.320\textwidth]{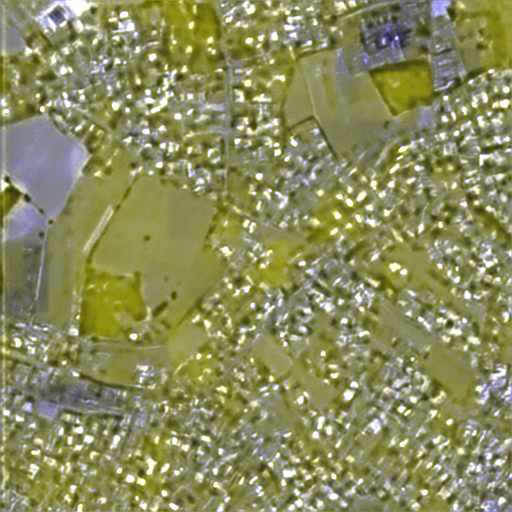}}
    \begin{tabular}{ccc}
        \makebox[0.30\textwidth]{\textbf{Low Resolution}} &
        \makebox[0.30\textwidth]{\textbf{GLP \cite{vivone2018full}}} &
        \makebox[0.30\textwidth]{\textbf{GLP Fusion Network (ours)}}
    \end{tabular}

    \subfigure{\includegraphics[width=0.320\textwidth]{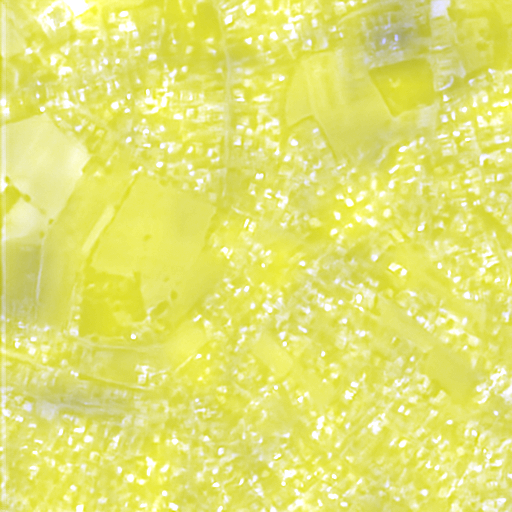}}
    \hfill
    \subfigure{\includegraphics[width=0.320\textwidth]{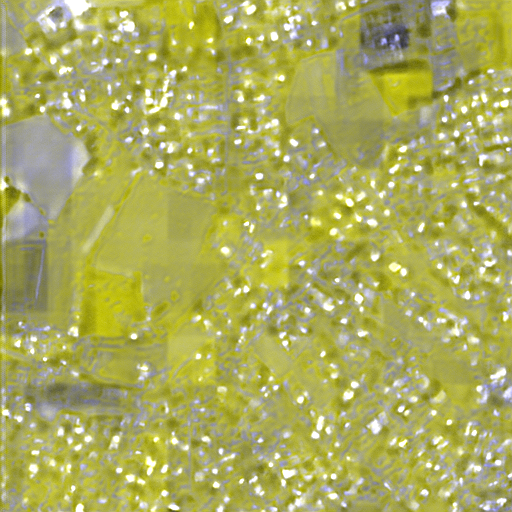}}
    \hfill
    \subfigure{\includegraphics[width=0.320\textwidth]{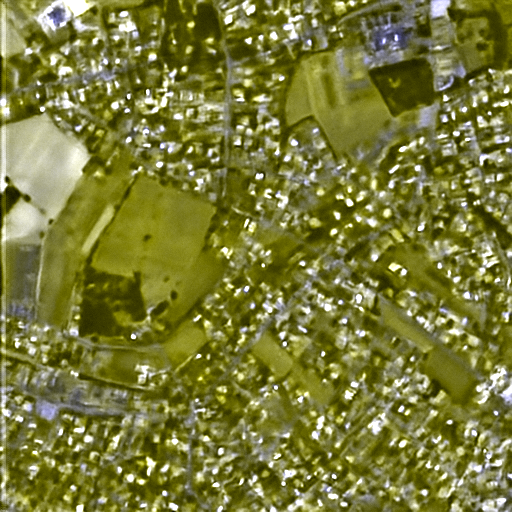}}
    \begin{tabular}{ccc}
        \makebox[0.30\textwidth]{\textbf{IHS}} &
        \makebox[0.30\textwidth]{\textbf{Fusion Network \cite{deng2022machine}}} &
        \makebox[0.30\textwidth]{\textbf{Gram Schmidt \cite{dalla2015global}}}
    \end{tabular}

    \caption{\add{Fusion output comparisons across 60\,m atmospheric bands. The figure show the Low Resolution input, Generalized laplacian pyramid (GLP) baseline method \cite{vivone2018full}, our GLP inspired neural network, intensity-hue-saturation (IHS) transform, and Fusion network baseline \cite{deng2022machine} and  Gram Schmidt \cite{dalla2015global} method. }}

    \label{fig:fusion3}
\end{figure*}

Fig.~\ref{fig:fusion1} and ~\ref{fig:fusion3} show results that are supplementary to the ones in the main paper. They evaluate the proposed learnable fusion model. Here, we only use the harmonised NAIP-based SR model to create the RGB super-resolved image that is further used for fusion.  The columns represent the Low Resolution input, Generalized laplacian pyramid (GLP) baseline method \cite{vivone2018full}, our GLP inspired neural network, intensity-hue-saturation (IHS) transform, and Fusion network baseline \cite{deng2022machine} and Gram Schmidt \cite{dalla2015global} method applied to 60 m bands and visualized in RGB, where the red and green channels use the Water Vapour band (B10) and the blue channel uses the Coastal Aerosol band (B1). Compared to the baselines, which often distort the structure and exaggerate spectral contrast, our GLP inspired fusion networks output is visually more consistent with the low-resolution signal while enhancing spatial detail.

\end{document}